\documentclass[journal]{IEEEtran}

\usepackage{graphicx}
\usepackage{citesort}
\usepackage{amssymb}
\usepackage{amsfonts}
\usepackage{amsmath}
\usepackage{epsfig}
\usepackage{color}
\usepackage{fancybox}
\usepackage{textcomp}
\usepackage{multirow}
\usepackage{setspa ce}
\usepackage{psfrag}
\usepackage[ruled,vlined]{algorithm2e}

\usepackage{booktabs}

\usepackage{mathrsfs}

\usepackage{tikz}

    \def\Complex{{\rm\rule[.23ex]{.03em}{1.1ex}\kern-.3em{C}}}

    \newcommand{\be}{\begin{equation}} \newcommand{\ee}{\end{equation}}
    \newcommand{\bea}{\begin{eqnarray}} \newcommand{\eea}{\end{eqnarray}}
    \newcommand{\benum}{\begin{enumerate}} \newcommand{\eenum}{\end{enumerate}}

    \newcommand{\qb}{{\bf b}}
    \newcommand{\qc}{{\bf c}}

    \newcommand{\qh}{{\bf h}}

    \newcommand{\qn}{{\bf n}}

    \newcommand{\qr}{{\bf r}}
    \newcommand{\qs}{{\bf s}}

    \newcommand{\qv}{{\bf v}}
    \newcommand{\qw}{{\bf w}}
    \newcommand{\qx}{{\bf x}}
    \newcommand{\qy}{{\bf y}}
    \newcommand{\qz}{{\bf z}}

    \newcommand{\qA}{{\bf A}}

    \newcommand{\qI}{{\bf I}}

    \newcommand{\qU}{{\bf U}}
    \newcommand{\qV}{{\bf V}}
    \newcommand{\qW}{{\bf W}}

    \newcommand{\qzero}{{\bf 0}}

    \newcommand{\qbeta}{{\boldsymbol \beta}}

    \newcommand{\qmu}{{\boldsymbol \mu}}

    \newcommand{\qtheta}{{\boldsymbol \theta}}
    \newcommand{\qsigma}{{\boldsymbol \sigma}}

    \newcommand{\bbR}{{\mathbb R}}
    \newcommand{\bbC}{{\mathbb C}}

    \newcommand{\calN}{{\mathcal N}}

    \newcommand{\diag}{{\tt diag}}
    
    \newcommand{\tr}{{\tt tr}}

\begin{document}

\title{Phase Retrieval using Expectation Consistent Signal Recovery Algorithm based on Hypernetwork}

\author{~Chang-Jen~Wang, Chao-Kai~Wen, \IEEEmembership{Senior Member,~IEEE}, Shang-Ho~(Lawrence)~Tsai, \IEEEmembership{Senior Member,~IEEE}, \\Shi~Jin, \IEEEmembership{Senior Member,~IEEE}, and Geoffrey Ye Li, \IEEEmembership{Fellow,~IEEE}

\thanks{C.-J.~Wang  and C.-K. Wen are with the Institute of Communications Engineering, National Sun Yat-sen University, Taiwan. E-mail: dkman0988@gmail.com, chaokai.wen@mail.nsysu.edu.tw}
\thanks{S.-H. Tsai is with the Department of Electrical Engineering, National Chiao Tung University, Hsinchu, Taiwan. E-mail: shanghot@mail.nctu.edu.tw}
\thanks{S. Jin is with the National Mobile Communications Research Laboratory, Southeast University, Nanjing, China. E-mail: jinshi@seu.edu.cn}
\thanks{G. Y. Li is with the Department of Electrical and Electronic Engineering, Imperial Colledge London, London, UK. E-mail: geoffrey.li@imperial.ac.uk}
}

\maketitle

\begin{abstract}
Phase retrieval (PR) is an important component in modern computational imaging systems. Many algorithms have been developed over the past half-century.
Recent advances in deep learning have introduced new possibilities for a robust and fast PR. An emerging technique called deep unfolding provides a systematic connection between
conventional model-based iterative algorithms and modern data-based deep learning. Unfolded algorithms, which are powered by data learning, have shown remarkable performance and convergence speed improvement over original algorithms. 
Despite their potential, most existing unfolded algorithms are strictly confined to a fixed number of iterations when layer-dependent parameters are used.
In this study, we develop a novel framework for deep unfolding to overcome existing limitations.
Our development is based on an unfolded generalized expectation consistent signal recovery (GEC-SR) algorithm, wherein damping factors are left for data-driven learning.
In particular, we introduce a hypernetwork to generate the damping factors for GEC-SR. Instead of learning a set of optimal damping factors directly, the hypernetwork learns how to generate the optimal damping factors according to the clinical settings, thereby ensuring its adaptivity to different scenarios.
To enable the hypernetwork to adapt to varying layer numbers, we use a recurrent architecture to develop a dynamic hypernetwork that generates a damping factor that can vary online across layers.
We also exploit a self-attention mechanism to enhance the robustness of the hypernetwork.
Extensive experiments show that the proposed algorithm outperforms existing ones in terms of convergence speed and accuracy and still works well under very harsh settings, even under which many classical PR algorithms are unstable.
\end{abstract}


\section*{I. Introduction}
The problem of reconstructing a complex vector from its linear transform magnitude, which is known as phase retrieval (PR), arises in many imaging applications \cite{Bian-OE15,Millane-JOSA90,Misell-JPD73,Shechtman-SPM15}. Conventionally, PR algorithms use an iterative process between the measurement and the target domain, such as Gerchberg-Saxton \cite{Freund-PA90} and Fienup \cite{Fienup-OL78} algorithms, to recover the phase from the magnitude. They usually require many iterations to converge. 
Thus, an approach solved the original non-convex PR problem by using semidefinite programming (e.g., PhaseLift \cite{Candes-CPAM13}). 
To reduce the complexity of PhaseLift, PhaseMax \cite{Goldstein-TIT18} is proposed to avoid lifting the convex relaxation formulations. Then, PhaseLamp \cite{Dhifallah-Allerton17} proposed to solve a sequence of PhaseMax problems.
Another approach solved the original non-convex PR problem with a two-stage approach, such as Wirtinger Flow (WF) \cite{Candes-TIT15}, reshaped WF (RWF) \cite{Zhang-17JMLR}, reweighted WF \cite{Yuan-Applie17}, truncated amplitude flow (TAF) \cite{Wang-18TIT}, reweighted amplitude flow (RAF) \cite{Wang-18TSP}.
Recent popular trend is to use Bayesian frameworks with message-passing, such as prGAMP \cite{Schniter-TSP15,Ma-19TIT}, prSAMP \cite{Rajaei-IPOL17}, prVAMP \cite{Sharma-TCI19}, and generalized expectation consistent signal recovery (GEC-SR) \cite{He-ISIT17,Wang-2020}. Among the algorithms, those on the basis of the Bayesian frameworks have demonstrated better performance because these methods take advantage of prior information on the signal and specific magnitude-only models \cite{Sharma-TCI19,Wang-2020}.
 
Recent advances in artificial intelligence (AI) technology have opened a new possibility for robust and fast PR and can be categorized into two approaches: data-driven and model-driven approaches. For the data-driven approach, a deep neural network (DNN) is used as a black box to solve an application-specific mapping, and the mapping is learned from a large amount of data without exploiting mathematical description \cite{Kappeler-ICIP2017,Rivenson-2018,Hand-18ArXiv,Tayal-20ArXiv,Tayal-20NeurIPS}. Despite providing unprecedented performance gains, the future development and practical deployment of such DNNs are hindered by their black-box nature, such as lack of interpretability, need for huge training sets, and unpredictability in new tasks.
For the model-driven approach, a network topology is constructed on the basis of domain knowledge.
For example, deep unfolding (or unrolling) \cite{Metzler-arXiv2018,He-19WC,Gregor-10ICML,He-19WC,He-20TSP,Goutay-arXiv2020,Monga-19ArXiV,Adler-TMI2018,Li-TCI2020,Diamond-arXiv2017,Metzler-ICNPS2017,Ryu-arXiv19,Zhang-2020,Hyder-20ECCV,Wei-20ICML,Ito-19TSP,Naimipour-Arxiv20,Wang-SPL20}
provides a concrete and systematic connection between DNNs and iterative algorithms. Deep unfolding basically unwraps an iterative algorithm into multiple layers by using an accessible algorithm as an initialization step, and replacing partial functions with DNNs. According to the level of involved learnable parameters (from high to low), we may broadly classify the whole spectrum of unfolded algorithms \cite{Monga-19ArXiV,Adler-TMI2018,Li-TCI2020,Diamond-arXiv2017,Metzler-ICNPS2017,Ryu-arXiv19,Zhang-2020,Hyder-20ECCV,Wei-20ICML,Ito-19TSP,Naimipour-Arxiv20,Wang-SPL20}
into three types.
 
\emph{Learning a proximal operator}
\cite{Monga-19ArXiV,Adler-TMI2018,Li-TCI2020}. In the first type, a DNN is used to replace all instances of proximal operations (or inverse functions) in the iteration process.
Such networks perform poorly when the forward model deviates from the training distributions.

\emph{Learning a denoiser}
\cite{Diamond-arXiv2017,Metzler-ICNPS2017,Ryu-arXiv19,Zhang-2020,Hyder-20ECCV}. In the second type, a DNN is used to replace a denoiser (or named regularizer, image prior model) in the iterative process. The denoiser directly learned from data is promising for improving reconstruction performance. The well-known plug-and-play (PnP) ADMM \cite{Ryu-arXiv19,Zhang-2020} is an example of this type.

\emph{Learning a few adjustable variables} \cite{Wei-20ICML,Ito-19TSP,Naimipour-Arxiv20,Wang-SPL20}. In the third type, all the mathematical functions in the original iterative algorithm are kept while only a handful of adjustable variables are introduced to improve the convergence speed and performance of the original algorithm. Such type requires much fewer parameters than the previous types, thereby reducing the demand for training data and training time. 

\textbf{Motivation and Contribution.}
With deep unfolding, designers can select different types according to their prior knowledge about the forward models and the reconstruction signals.
A high-level idea behind this selection is that when the forward model is sufficiently explicit, the use of a small network with few parameters is sufficient to leverage the data-driven learning ability
and mitigate the convergence and initial conditions of the original algorithms.
Despite the potential advantages, existing unfolded algorithms have some limitations.
First, the learning parameters of the existing unfolded algorithms are
trained for a specific task of image recovery. Retraining the parameters is often needed in a
clinical setting, where different forward models (e.g., measurement distribution and size, and noise level) may be used; otherwise, the stability and optimality of the learned algorithm will be lost. Second, given the dynamical nature of the application scenarios, the iteration numbers of the algorithm should be adaptive.
However, most existing unfolding models with layer-dependent parameters are strictly confined to a fixed number of iterations to ensure that the number of free parameters is fixed, thereby lacking flexibility.

To overcome the two limitations, we attempt to develop a novel framework to build deep unfolding.
As a practice for PR, our development is based on an unfolded Bayesian algorithm called GEC-SR-Net \cite{Wang-SPL20}, which has two enhancements that lead to significant improvement in real-world applications.
In GEC-SR-Net, all functions are computed as a solution to the data-free Bayesian estimation problem, and only damping factors are left for data-driven learning.
GEC-SR-Net has already shown excellent accuracy and speed over many existing PR algorithms \cite{Sharma-TCI19,Wang-2020} while possessing the mentioned limitations.
Our framework can be distilled into three strategies as follows.
\begin{itemize}
\item \emph{Hypernetwork unit for adaptability.} We establish a new deep unfolding architecture by introducing a hypernetwork that is used as another network to \emph{generate} the parameters for the original unfolded algorithm. In contrast to GEC-SR-Net, whose damping factors are directly learned from a specific task for image recovery, we introduce a hypernetwork called GEC-SR-HyperNet to generate the damping factors for GEC-SR-Net. The hypernetwork takes a set of inputs that contain information about the forward models and generates the damping factors for GEC-SR-Net as its outputs, thereby ensuring adaptivity to different scenarios.

\item \emph{Dynamic architecture for flexibility.}  The number of damping factors in GEC-SR-Net is fixed, and these learned damping factors are customized for the fixed iteration architecture. To make the hypernetwork work with varying layer numbers, we use a recurrent neural network (RNN) \cite{LSTM,GRU} to develop a dynamic hypernetwork that dynamically generates a damping factor that can vary online across layers. In particular, we train the dynamic hypernetwork to serve as a controller that initializes the damping factors according to forward models and can adaptively adjust the damping factors \emph{online} by observing the convergence state of each iteration to prevent divergence or to accelerate speed.

\item \emph{Attention mechanism for robustness.} The hypernetwork intends to generate a set of optimal damping factors for GEC-SR-Net given a proper set of inputs. However, the proper inputs to generate damping factors should be changed dynamically under different scenarios. That is, the hypernetwork should pay attention to different input features under various scenarios. To this end, we introduce an attention mechanism called self-attention \cite{Attention} to compute a representation of inputs. The self-attention function can relate the different positions of an input vector to compute a new representation of the input vector under different scenarios. Findings show that the hypernetwork with attention enables superior robustness over various scenarios even when the measurement sizes and distributions are completely different in the clinical setting.
\end{itemize}

In the most relevant work \cite{Wei-20ICML}, a policy network that is obtained via deep reinforcement learning automatically determines the
internal parameters of PnP-ADMM, including the penalty parameter, the denoising strength, and the terminal time.
Thus, hypernetwork and the policy network share the same spirit. Both can be interpreted as smart controllers for iterative algorithms.
The hypernetwork should have fewer parameters and must be easier to train than the policy network.
 
{\bf Notations.} $\mathcal{N}_{\mathbb{C}}(x; \mu,v)$ is the complex Gaussian distribution of dummy variable $x$ with mean $\mu$ and variance $v$.
We define vector-vector multiplication and division as their component-wise vector multiplication and division, respectively.
Notably, in this work, the standard operators $\cdot$, $/$, $(\cdot)^2$, $(\cdot)^{-1}$, and $|\cdot|$ on a vector are all defined as element-wise operators.
${\bf 1}$ denotes a vector of ones.

\section*{II. Problem Setup and Algorithm Framework}

\begin{figure*}[ht]
\begin{center}
\resizebox{7.0 in}{!}{
\includegraphics*{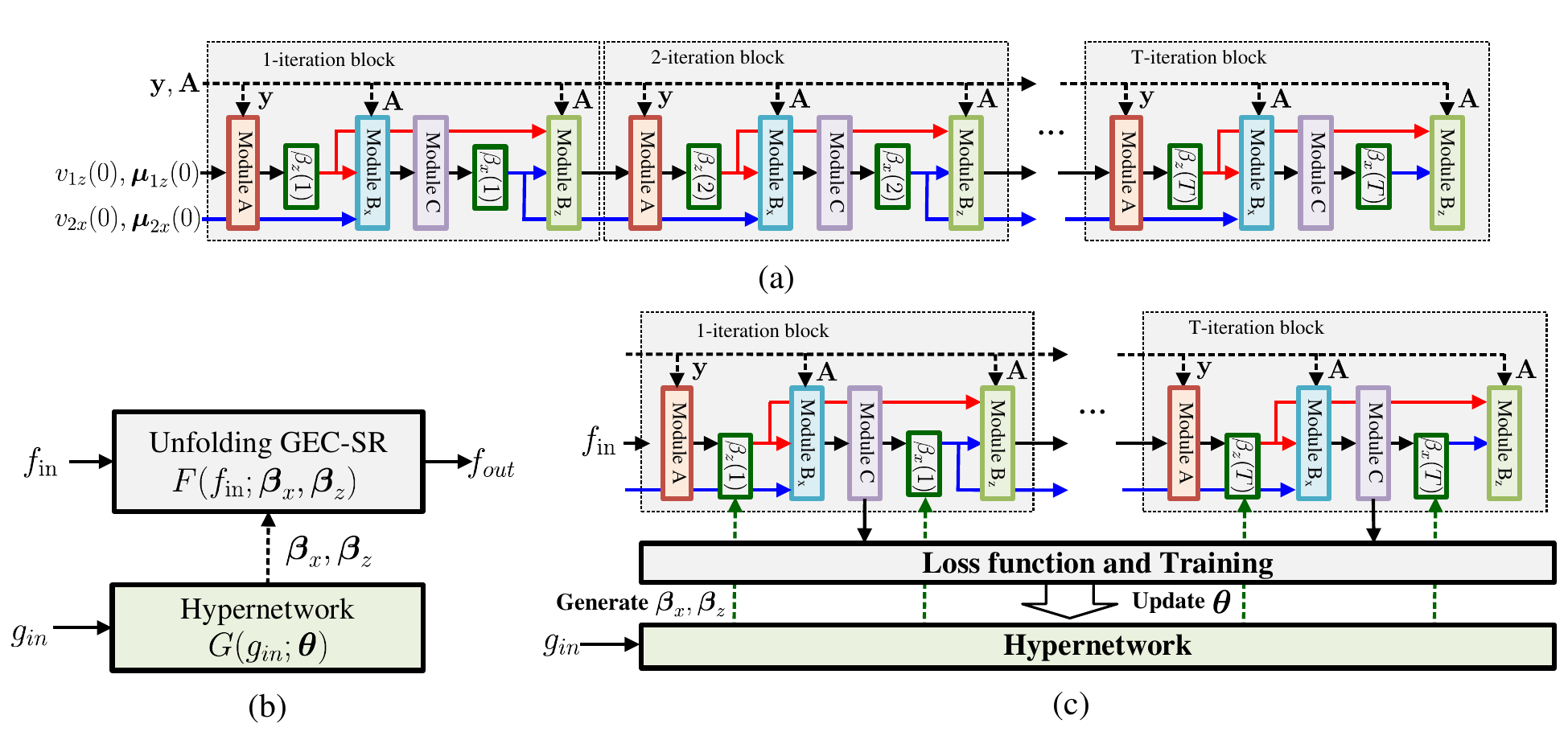} }
\caption{Block diagrams of (a) GEC-SR and (b) the structure for neural network with hypernetwork. (c) Data-driving tuning is based on an unfolding GEC-SR with hypernetwork,
where the blue and red arrows indicate the estimates of $z$ and $x$ through the damping operation, respectively.}\label{fig:Fig1-deGEC-diag}
\end{center}\vspace{-0.3cm}
\end{figure*}

We consider the PR problem in which unknown signal $\qx \in \mathbb{C}^{N}$ is observed via phase-less measurements, $\qy \in \mathbb{R}^{M}_{+}$, which are expressed as 
\begin{equation}\label{eq: sysModel}
\qy = {\tt Q}(\qA\qx+\qn) ,
\end{equation}
where $\qA \in \mathbb{C}^{M\times N}$ is the linear transform matrix, $\qn \in \mathbb{C}^{M}$ is a noise vector, and operator ${{\tt Q}(\cdot)}$ takes the element-wise ${|\cdot|}$.
PR aims to recover signal $\qx$ from measurements $\qy$.

In this work, we consider the case where signal ${\qx \in \mathbb{C}^{N}}$ is generated following a prior distribution $p(\qx)$, and noise $\qn$ is the standard circularly symmetric Gaussian random vector.\footnote{The standard complex Gaussian random vector is a complex random vector whose components are independent complex random variables with real and imaginary parts being independent normally distributed random variables with mean zero and variance $1/2$. Formally, we write $\qn \sim \calN_{\bbC}(\qn; \qzero, \qI)$.} We assume that transform matrix $\qA$, prior distribution $p(\qx)$, and the noise distribution, are known. Therefore, the posterior distribution can be computed by
\begin{equation} \label{eq: posterior}
 p(\qx|\qy) = \frac{p(\qy|\qx) p(\qx) }{p(\qy)},
\end{equation}
where $p(\qy|\qx)$ is the likelihood of measurements $\qy$ given that $\qx$ is the true underlying signal, and $p(\qy) = \int p(\qy|\qx) p(\qx) d\qx $ is the marginal distribution.
Thus, solving the Bayes estimate can recover signal $\qx$ by using
\begin{equation}\label{eq: mmse}
 \hat{\qx} = \int \qx \, p(\qx|\qy) d\qx.
\end{equation}
In principle, the Bayes estimator can be used to solve the reconstruction problem, and the
direct computation of \eqref{eq: mmse} is generally intractable when the distribution of the signal cannot make the posterior distribution in close form.

In the last 10 years, various methodologies have been developed to solve the Bayes estimation problem.
Most notable approaches consider approximating $p(\qx|\qy)$ with a tractable density $q(\qx)$ up to some convenient factorizations of $p(\qx|\qy)$.
A message-passing algorithm then works by iteratively minimizing the Kullback-Leibler divergence from $p(\qx|\qy)$ to $q(\qx)$ \cite{Opper-NC2000,Minka-PSCUAI01}. The approximation can be made in many ways. GEC-SR \cite{Fletcher-ISIT16,He-ISIT17} uses expectation consistent (or moment matching) approximation
and has demonstrated excellent performance in terms of reconstruction accuracy, convergence speed, and robustness.

GEC-SR belongs to a group of message-passing algorithms.
For ease of developing an unfolding framework, we introduce GEC-SR in a \emph{modular-based} manner \cite{Meng-18SPL,Zhu-SPL19,Xue-18ArXiv} rather than a message-passing manner.
Let us first define a hidden vector ${\qz = \qA \qx}$.
From \eqref{eq: sysModel}, measurements $\qy$ are related to $\qx$ following the procedure
\begin{equation} \label{eq:forward_direction}
 \buildrel{p_{x}}\over\hookrightarrow  \qx \buildrel{\qA}\over\longrightarrow \qz \buildrel{{\tt Q}}\over\longrightarrow \qy,
\end{equation}
where operator $ \buildrel{p_{x}}\over\hookrightarrow $ represents that $\qx$ is generated by following a prior distribution $p(\qx)$.
Similarly, we use operator $\buildrel{p^{-1}}\over\hookleftarrow$ to denote that $\qx$ is restricted to the support of distribution $p(\qx)$,
and we use operators ${\tt Q}^{-1}$ and ${\tt A}^{-1}$ to denote the inverses of ${\tt Q}$ and the linear transform, respectively.
The specific choice of these inverse operators depends on particular approximation approaches. For example,
common choices of ${\tt A}^{-1}$ include the adjoint $\qA^{H}$ or pseudoinverse $\qA^{-1}$ but are not limited to these \cite{Ma-SPL15,MA-NC2017}.
With \eqref{eq:forward_direction}, the inference of $\qx$ follows a reverse procedure, that is,
\begin{equation}
 \qy \buildrel{{\tt Q}^{-1}}\over\longrightarrow \qz \buildrel{A^{-1}}\over\longrightarrow \qx \buildrel{p_{x}^{-1}}\over\hookleftarrow ,
\end{equation}
which is achieved through Modules ${\rm A}$, ${\rm B}$, and ${\rm C}$ (Fig. \ref{fig:Fig1-deGEC-diag}(a)).

\begin{itemize}
\item Module ${\rm A}$ computes (mean and variance) the estimates of $\qz$ by using measurements $\qy$ and the prior knowledge of $\qz$ (which is from Module ${\rm B}_{\rm z}$).
Module ${\rm A}$ can be interpreted as a \emph{phase reconstructor} because $\qz$ is estimated through a PR process from the phase-less measurements, $\qy$.

\item Module ${\rm B}$ provides estimates of $(\qz,\qx)$ according to the linear transformation of $\qz = \qA\qx$.
The inputs of Module ${\rm B}$ are $(\qz,\qx)$, wherein the priors of $\qz$ and $\qx$ are from Modules ${\rm A}$ and ${\rm C}$, respectively.
The output of Module ${\rm B}$ is either $\qx$ or $\qz$, depending on its processing direction (feed-forward or feedback).
We use Modules ${\rm B}_{\rm x}$ and ${\rm B}_{\rm z}$ to indicate the outputs of Module ${\rm B}$ being $\qx$ and $\qz$, respectively.
In Module ${\rm B}$, $(\qz,\qx)$ are assumed to be Gaussian distributed to facilitate estimation, and their true prior distributions are not used. Module ${\rm B}$ can be interpreted as a \emph{linear reconstructor}.

\item Module ${\rm C}$ is similar to Module ${\rm A}$ but now used for signal $\qx$.
In particular, Module ${\rm C}$ computes (mean and variance) the estimates of $\qx$ by using the estimates from Module ${\rm B}_{\rm x}$
and the prior knowledge $p(\qx)$. Module ${\rm C}$ can be interpreted as a \emph{denoiser} (or regularizer) because $\qx$ is estimated on the basis of its true prior distribution $p(\qx)$.
\end{itemize}

Modules ${\rm A}$, ${\rm B}$, and ${\rm C}$ can be generally understood as reconstructions of $\qz$ or $\qx$ under different prior knowledge.
Proceeding through Modules A and B$_{\rm x}$, the first estimate of $\qx$ is obtained in Module ${\rm C}$ via the feed-forward direction procedure. Then, we reconstruct back to $\qy$ in a feedback direction procedure.
That is, given $\qx$ from Module ${\rm C}$, Module ${\rm B}_{\rm z}$ is used to provide the estimates of $\qz$. We
then use Module ${\rm A}$ to refine the estimates of $\qz$ by combining the phase-less measurements $\qy$.
The three modules are executed iteratively in the manner of
\begin{equation} \label{eq:IterativeManner}
\underbrace{{\rm A} {\rightarrow} {\rm B}_{\rm x} {\rightarrow}
{\rm C} {\rightarrow} {\rm B}_{\rm z} {\rightarrow}}_{\rm iteration\, 1}
\underbrace{{\rm A} {\rightarrow} {\rm B}_{\rm x} {\rightarrow} {\rm C} {\rightarrow} {\rm B}_{\rm z}}_{\rm iteration\, 2}
{\rightarrow} \cdots
\end{equation}
until convergence.

GEC-SR can be applied to solve other inverse problems, such as quantization, as long as the forward operator ${\tt Q}(\cdot)$ is defined. Therefore, our developments in the following sections are also applicable to other inverse problems.

\section*{III. Unfolding and Hypernetwork}

To develop deep unfolding based on hypernetwork, we first introduce the unfolded GEC-SR algorithm \cite{Wang-2020}
in Section III.A. Next, we describe the construction of a hypernetwork to generating the parameters of the unfolded GEC-SR in Section III.B. Finally, several architectures of the hypernetwork that can improve
the flexibility and robustness of the unfolded GEC-SR are described in Section III.C.

\subsection*{A. Unfolded GEC-SR}

Inspired by the modular-based iterative manner in \eqref{eq:IterativeManner}, we introduce the unfolding method in \cite{Diamond-arXiv2017,Monga-19ArXiV,He-19WC} and use it to develop multiple layers of the reconstruction network, namely,  GEC-SR-Net \cite{Wang-SPL20} (Fig. \ref{fig:Fig1-deGEC-diag}(a)). The modules in each iteration step unfold into a single network layer.
The number of layers is the same as the number of iterations.
We can apply statistical inference on each module to achieve its particular reconstruction purpose, which is referred to as model-based.
We can also turn a certain module into a DNN trained by using real datasets, which is referred to data-based.
For example, Module ${\rm C}$ can be replaced by a DNN that acts as a denoiser for the reconstruction, such as the plug-and-play recover \cite{Ryu-arXiv19,Zhang-2020}.

In GEC-SR-Net, the model-based modules that are the same as those in the original GEC-SR are inherited while introducing free parameters in damping operations after Modules ${\rm A}$ and ${\rm C}$.
The damping factors can be learned by real datasets.
This design strategy is based on the argument that model-based modules are mathematically well developed with few uncertainties.
Keeping the modules in a model-based manner enables the whole reconstruction network to still work even under untrained scenarios.
However, updating these modules in a simple manner often yields some convergence
issues, such as oscillation and non-positive definiteness in approximated moments.

Damping is applied to the updates to fix this problem. Conservative damping factors result in intolerably slow convergence, whereas aggressive damping factors result in divergence.
The optimum damping factors vary from scenario to scenario, and the tuning process is mostly customized.
Therefore, we leverage deep learning to generate a suitable damping factor for each iteration to reduce the number of iterations and further increase the reconstruction accuracy.
In summary, we leverage the partial knowledge of the system model in \eqref{eq: sysModel}
to perform some approximate inversion of the measurement process while using training data
to \emph{learn to remove the manual tuning and compensate for the mismatch in model approximation}.

Before proceeding, we briefly introduce each module in GEC-SR.
Generally, Modules ${\rm A}$, ${\rm B}_{\rm x}$, ${\rm C}$, and ${\rm B}_{\rm z}$ comprise 1) the Bayesian estimation operation and 2) an extrinsic (or debias) operation.
To describe the above operations in a generic form, we let ``${\sf a}$'' and ``${\sf b}$,'' either same or different, be the random variables of either $\qx$ or $\qz$.

{\bf Bayesian Estimation Operation}---Given the mean and variance $(\qmu_{1{\sf a}}, v_{1{\sf a}})$ from a previous module, we define the posterior distribution
\begin{equation} \label{eq:posterWithGaussian}
   {f({\sf a}|{\sf b}) = \frac{f({\sf b}|{\sf a}) \calN_{\bbC}({\sf a};\qmu_{1{\sf a}}, v_{1{\sf a}} )}{{\cal Z}}},
\end{equation}
where $f({\sf b}|{\sf a})$ is a likelihood function of ${\sf b}$ condition on ${\sf a}$, and ${{\cal Z} = \int f({\sf b}|{\sf a}) \calN_{\bbC}({\sf a};\qmu_{1{\sf a}},v_{1{\sf a}}) d{\sf a}}$ performs normalization.
The Bayesian (or posterior) estimate of the mean and variance of ${\sf b}$ are obtained by
\begin{align}
\widehat{\qmu}_{1{\sf b}} &= \int {\sf b} f({\sf a}|{\sf b}) d{\sf b}, \label{eq:exp_mean} \\
\widehat{\qv}_{1{\sf b}} &= \int |{\sf b}|^2  f({\sf a}|{\sf b}) d{\sf b} - |\widehat{\qmu}_{1{\sf b}}|^2. \label{eq:exp_var}
\end{align}
The posterior variance, $\widehat{\qv}_{1{\sf b}}$, is the corresponding mean-squared error (MSE) of the posterior mean $\widehat{\qmu}_{1{\sf b}}$. Notice that
the posterior variance, $\widehat{\qv}_{1{\sf b}}$, is in vector form because the calculation in \eqref{eq:exp_var} is performed  component-wise.
In this work, we always consider feeding the average of $\widehat{\qv}_{1{\sf b}}$ to the next module, that is
\begin{equation} \label{eq:avg_v1b}
    \widehat{v}_{1{\sf b}} =  \frac{{\bf 1}^T \widehat{\qv}_{1{\sf b}}}{  N_{\widehat{v}_{1{\sf b}}}},
\end{equation}
where $N_{\widehat{v}_{1{\sf b}}}$ denotes the length of $\widehat{\qv}_{1{\sf b}}$.
For ease of notation, we express \eqref{eq:exp_mean} and \eqref{eq:avg_v1b} in pairs as
\begin{equation} \label{eq:exp_pair}
 (\widehat{\qmu}_{1{\sf b}}, \widehat{v}_{1{\sf b}}) = \mathbb{E} {\left\{{\sf b}|\qmu_{1{\sf a}}, v_{1{\sf a}}; f({\sf b}|{\sf a}) \right\}}.
\end{equation}

{\bf Extrinsic Operation}---For efficient message passing, each module only passes extrinsic messages (or unbiased estimates) \cite{Brink-99,Xue-19ArXiv} to the next module rather than the posterior estimates.
We use subscripts $(\cdot)_{1}$ and $(\cdot)_{2}$ to represent input priors and extrinsic messages, respectively. If the input priors are denoted by $(\cdot)_{1}$, then their corresponding extrinsic messages are denoted by  $(\cdot)_{2}$, and vice versa.
The extrinsic mean and variance of $(\widehat{\qmu}_{1{\sf b}}, \widehat{v}_{1{\sf b}})$
are calculated by excluding the prior mean $\qmu_{1{\sf a}}$ and variance $v_{1{\sf a}}$ and are given as\footnote{We assume that posterior estimate of ${\sf b}$ is acted as Gaussian with mean $\widehat{\qmu}_{1{\sf b}}$ and variance $\widehat{\qv}_{1{\sf b}}$, and the prior of ${\sf b}$ is also acted as Gaussian with mean $\widehat{\qmu}_{2{\sf b}}$ and variance $\widehat{\qv}_{2{\sf b}}$. Hence, we can obtain the extrinsic messages as Gaussian using the following Gaussian product rule. A product of two Gaussians with the same argument but different means
and variances has the following formula:
\begin{equation}
 \calN(x; \mu_1, v_1) \calN(x; \mu_2, v_2) = \calN(x; \mu, v), \notag
\end{equation}
where $\mu = v (\mu_1/v_1+\mu_2/v_2)$ and $v = (1/v_1 + 1/v_2)^{-1} $.
}
\begin{subequations}\label{eq:extMeanAndVar}
\begin{align}
     \qmu_{2{\sf a}} &= v_{2{\sf a}} {\left(\frac{\widehat{\qmu}_{1{\sf b}}}{\widehat{v}_{1{\sf b}}} - \frac{\qmu_{1{\sf a}}}{v_{1{\sf a}}}\right)},\\
     v_{2{\sf a}} &= {\left(\frac{1}{\widehat{v}_{1{\sf b}}} - \frac{1}{v_{1{\sf a}}}\right)}^{-1}.
\end{align}
\end{subequations}
Similarly, we express the extrinsic estimates in pairs as
\begin{equation} \label{eq:ext_def}
 (\qmu_{2{\sf a}}, v_{2{\sf a}}) \Leftarrow  {\left(\widehat{\qmu}_{1{\sf b}},\widehat{v}_{1{\sf b}}\right)} \backslash {\left(\qmu_{1{\sf a}}, v_{1{\sf a}}\right)}.
\end{equation}

In summary, given the mean and variance $(\qmu_{1{\sf a}}, v_{1{\sf a}})$ from the previous module as the inputs, the subsequent module performs Bayesian
estimations $(\widehat{\qmu}_{1{\sf b}},\widehat{v}_{1{\sf b}})$ by using \eqref{eq:exp_pair} and then outputs extrinsic messages $(\qmu_{2{\sf a}}, v_{2{\sf a}})$ by using \eqref{eq:ext_def}.
The Bayesian estimations in \eqref{eq:exp_pair} can be expressed explicitly, and details can be found in \cite{Wang-2020}.

Except for Modules ${\rm A}$, ${\rm B}_{\rm x}$, ${\rm C}$, and ${\rm B}_{\rm z}$, a single network-layer also comprises the damping operations.
As shown in Fig. 1(c), Modules ${\rm A}$ and ${\rm C}$ are cascaded with a damping operation.

{\bf Damping Operation}---We use ``$\qmu(t)$'' and ``$v(t)$'' to represent the output of either Module ${\rm A}$ or ${\rm C}$, respectively, in the $t$-th layer of the network.
Next, given an initialization $(\qmu(0), v(0))$, the damped update is carried out by
\begin{equation}
 {\tt Damp}{\left(\qmu(t), v(t); \beta(t)\right)} = {\left[
 \begin{array}{l}
 \beta(t) \qmu(t-1)+(1-\beta(t)) \qmu(t)  \\
 \beta(t) v(t-1)+(1-\beta(t)) v(t)
 \end{array}
 \right]}, \label{eq:damping_function}
\end{equation}
for $t = 1, \ldots, T$, where $\beta(t)\in [0, \, 1]$ is the damping factor.
The damping factors can be layer-dependent.
For example, a damping factor is often started off with a small value and gradually increased with the iterations.
In addition, the damping factors for Module ${\rm A}$ or ${\rm C}$ can be different.
We denote the damping factors for Modules ${\rm A}$ and ${\rm C}$ by $\beta_{z}(t)$ and $\beta_{x}(t)$, respectively, while we simply denote it by $\beta(t)$ if $\beta_{z}(t) = \beta_{x}(t)$.

When the modules and damping operators are combined, the structure of GEC-SR-Net is depicted as follows:
\begin{multline} \label{eq:circular}
\underbrace{{\rm A} {\rightarrow} {\tt Damp}(\cdot ;\beta_{z}(1)) {\rightarrow} {\rm B}_{\rm x} {\rightarrow}
{\rm C} {\rightarrow} {\tt Damp}(\cdot;\beta_{x}(1)) {\rightarrow} {\rm B}_{\rm z}}_{\rm Layer\, 1}  {\rightarrow}
\\ \cdots {\rightarrow} \underbrace{{\rm A} {\rightarrow} {\tt Damp}(\cdot ;\beta_{z}(T)) {\rightarrow} {\rm B}_{\rm x} {\rightarrow}
{\rm C} {\rightarrow} {\tt Damp}(\cdot;\beta_{x}(T)) {\rightarrow} {\rm B}_{\rm z}  }_{{\rm Layer}\, T} .
\end{multline}
The subsequent layers follow the same architecture as Layer 1, and we totally take the $T$ layers.
The theoretical state evolutions (SEs) of GEC-SR for PR can be referred to \cite{Wang-2020}. We can simply substitute the damping factors to the SEs of GEC-SR
to obtain a quick performance prediction for GEC-SR-Net. However, the SEs are derived under the large system limits and under certain assumptions, which may not be satisfied in application scenarios. Thus, the theoretical SEs cannot be used to determine the damping factors in practice.
In GEC-SR-Net, $v(0),\qmu(0)$ are initialized by the spectral initializer as in \cite{Netrapalli-TSP15} that
often provides adequate initialization to any iterative PR algorithm. In addition, the damping factors $\qbeta_{x} = [\beta_{x}(1),\ldots,\beta_{x}(T)]$ and  $\qbeta_{z} = [\beta_{z}(1),\ldots,\beta_{z}(T)]$ are learned directly from datasets.

Specifically, training is performed on the basis of $L$ samples using the training data of the form $( \qy^{l}, \qA^{l}, \qx^{l} )$ for $l=1,2,\ldots, L$, where transform matrix $\qA^{l}$ and signal $\qx^{l}$ are randomly generated for each sample, and $\qy^{l}$ is obtained using \eqref{eq: sysModel}.
By feeding $(\qy^{l},\qA^{l})$ into the network, GEC-SR-Net generates $\widehat{\qmu}_{1x}^{l}(t)$ at the $t$-th layer and eventually outputs $\widehat{\qmu}_{1x}^{l}(T)$ at the $T$-th layer as the reconstructed signal.
In RP, the reconstructed signal can only be recovered up to a
global phase difference. Therefore, to quantify the quality of the reconstructed signal, the ambiguity of each estimate must be removed by
\begin{equation}
{\tt dis}\left(\qx^{l}, \widehat{\qmu}_{1x}^{l}(t)\right) = e^{j \phi^{l}(t)} \widehat{\qmu}_{1x}^{l}(t),
\end{equation}
where $\phi^{l}(t) =\angle ( (\widehat{\qmu}_{1x}^{l}(t))^{H}   \qx^{l} )$.
Therefore, the loss function is defined as
\begin{equation} \label{eq:LossFun}
\mathcal{L}(\qbeta_{x}, \qbeta_{z}) = \frac{1}{L} \sum_{l=1}^{L} \sum_{t=1}^{T} \left\| \qx^{l} - {\tt dis}\left(\qx^{l}, \widehat{\qmu}_{1x}^{l}(t)\right)\right\|_2^2.
\end{equation}
An optimizer is used to tune $(\qbeta_{x},\qbeta_{z})$ through back-propagation, which minimizes the loss function in \eqref{eq:LossFun} between the true signal, $\qx^{l}$, and the estimate, $\widehat{\qmu}_{1x}^{l}(t)$, of every layer.

\subsection*{B. Hypernetworks}

In GEC-SR-Net, the damping factors are directly learned from training data to minimize the MSE of the reconstruction at each layer (or iteration). The learned damping factors also compensate for mismatches in model approximation and thus provide improved performance than the original GEC-SR.
Despite the excellent performance, GEC-SR-Net has two disadvantages.

First, after training, the learned damping factors are fixed in testing and deployment.
When statistical properties of data used during training are similar to those used during testing, GEC-SR-Net can exhibit excellent performance. However, a change in the statistical properties of datasets in testing, such as the distribution of $\qA$ or the signal-to-noise ratio (SNR) level, would require the damping factors to retrained. Alternatively, we can learn the damping factors through various statistical datasets.
However, because the needs of the damping factors vary from scenario to scenario, the learned damping factors would be too conservative to find a set of damping factors that can perform well in various scenarios, thereby resulting in limited improvement in convergence speed. The convergence speed is compromised with robustness in GEC-SR-Net.

Second, to apply the loss function in \eqref{eq:LossFun} for obtaining a set of damping factors, the number of layers in GEC-SR-Net should be fixed. For practical applications, the iteration number should be adjustable under different scenarios. For example, in a certain scenario with poor convergence conditions, one should be able to increase the number of iterations dynamically to improve the signal reconstruction performance and vice versa. In this case, GEC-SR-Net has to retrain the damping factors for different scenarios.

To resolve the two problems, we leverage the idea from a hypernetwork \cite{HyperNet}: an approach of using one network to generate the parameters for another network.
In this subsection, we first describe the construction of a hypernetwork to generate the damping factors
for GEC-SR-Net in a general form. Different hypernetwork architectures are presented in the next subsection.

Let ${ f_{\rm out} = F( f_{\rm in}; \qbeta_{x}, \qbeta_{z}) }$ be the primary network (i.e., GEC-SR-Net), where $f_{\rm out} $ and $f_{\rm in} $ represent the output and input, respectively;
and $(\qbeta_{x}, \qbeta_{z})$ are the damping factors that will be generated from the hypernetwork.
The behavior of the primary network is the same as GEC-SR-Net.
Specifically, the input of the primary network is $f_{\rm in} = \{ \qy^{l}, \qA^{l}, v_{1z}(0),\qmu_{1z}(0), v_{2x}(0), \qmu_{2x}(0) \}$, where $v_{1z}(0),\qmu_{1z}(0)$ and $v_{2x}(0), \qmu_{2x}(0)$ are determined by a spectral initializer.
The output of the primary network corresponds to the reconstructed signals of the $T$ layers, that is, $f_{\rm out} = \{ \widehat{\qmu}_{1x}^{l}(1), \cdots, \widehat{\qmu}_{1x}^{l}(T) \}$.
Similarly, let
\begin{equation}
 g_{\rm out} =  G(g_{\rm in}; \qtheta)
\end{equation}
be the hypernetwork, where $g_{\rm out}$ and $g_{\rm in}$ are the output and input, respectively; and $\qtheta$ consists of the hypernetwork parameters.
The hypernetwork takes a set of inputs that contain information about the damping factors and generates the damping factors for GEC-SR-Net as its output (Fig. \ref{fig:Fig1-deGEC-diag}(b)). We will discuss the architecture and input of the hypernetwork in the next subsection.
An optimizer is used to learn $\qtheta$ by minimizing the loss function
\begin{equation} \label{eq:LossFun_Hyper}
\mathcal{L}( G(g_{\rm in}; \qtheta) ) = \frac{1}{L} \sum_{l=1}^{L} \sum_{t=1}^{T} \left\| \qx^{l} - {\tt dis}\left(\qx^{l}, \widehat{\qmu}_{1x}^{l}(t)\right)\right\|_2^2.
\end{equation}
After training the primary network together with the hypernetwork, we obtain the hypernetwork $G(g_{\rm in}; \qtheta)$ that can generate the optimal damping factors $(\qbeta_{x}, \qbeta_{z})$ given a set of inputs.
We call the whole network as GEC-SR-HyperNet, which can be regarded as a rule integration of the parameters of GEC-SR-Net.

Notably, hypernetworks in \cite{HyperNet} are used to reduce the number of networks parameters that use a small network to generate
the weights for a larger network. The hypernetworks of \cite{HyperNet} can be seen as imposing weight-sharing layers.
In this work, we use hypernetworks to enable GEC-SR-Net to adapt to various scenarios instead of introducing weight sharing.
By feeding various scenarios of samples, the hypernetwork learns how to generate the damping factors of GEC-SR-Net.
In particular, hypernetwork learns an integration of the trend to generate the damping factors of GEC-SR-Net for different scenarios.
As a result, the generated damping factors still exhibited good performance even when the test scenarios are mismatched with the training scenarios.

\begin{figure}
\begin{center}
\resizebox{3.6in}{!}{
\includegraphics*{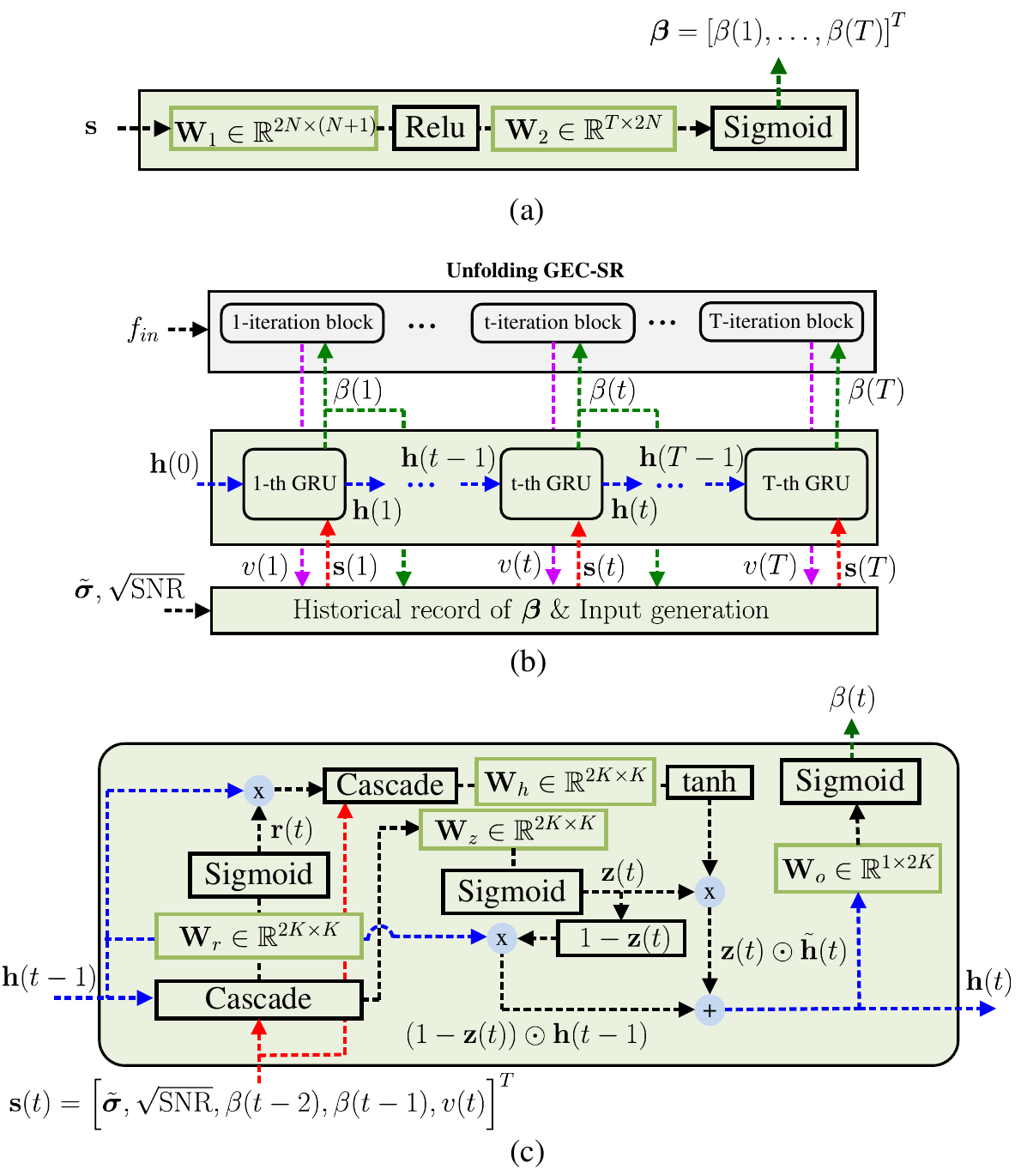} }
\caption{ Block diagrams of (a) GEC-SR-HyperNet and (b) GEC-SR-HyperGRU. (c) The $t$-th iteration GRU model has two kinds of input,
where the blue and red arrows indicate the hidden state information and external input information, respectively.
}\label{fig:Fig-HyperNet}
\end{center}\vspace{-0.6cm}
\end{figure}

\subsection*{C. Hypernetwork Architectures}

\subsubsection*{Static Hypernetwork}
The hypernetwork intends to generate a set of optimal damping factors for GEC-SR-Net when a proper set of inputs is given.
The architecture of this hypernetwork is motivated by the two observations.

First, empirical experiments \cite{Rangan-ISIT17,Wang-2020,He-ISIT17,Meng-19ACCESS} have shown that the convergence of GEC-SR is highly relevant to the distribution of the transform matrix, $\qA$. However, the dimension of $\qA$ is usually large. Thus, importing $\qA$ directly presents a computational burden to the hypernetwork.
In \cite{Wang-2020}, a state evolution of GEC-SR, which shows the MSE of
the reconstruction at each iteration can be a function of the singular values of $\qA$ and the noise level, is derived. Let
\begin{equation}
\qsigma = [\sigma_1, \ldots, \sigma_N]
\end{equation}
be the singular values of $\qA$.
Therefore, it is reasonable to let $\qsigma$ and the noise level be the input of the hypernetwork because the two parameters should be relevant to the convergence.

Second, the simulation results in \cite{Wang-SPL20} have shown that the performance of GEC-SR-Net, with $\qbeta_{x} = \qbeta_{z}$ or $\qbeta_{x} \neq \qbeta_{z}$, is comparable. Therefore, we can tie the two damping factors together (i.e., $\qbeta_{x} = \qbeta_{z}$) during the training process to reduce the complexity of the hypernetwork.
We simply use $\qbeta$ to denote the damping factors in this hypernetwork.

In \eqref{eq: sysModel}, we have normalized the noise to be standard Gaussian. In this case, the noise level is fixed to be one, and
thus the SNR is defined by
\begin{equation} \label{eq:snr_def}
    {\rm SNR}= \tr{(\qA\qA^H)} /M .
\end{equation}
Specifically, we have absorbed the noise level into $\qA$ by adjusting the scale of $\qA$ to satisfy different SNR specifications.
In this way, the dynamic range of $\qsigma$ should be very large.
To make the network robust to the wide dynamic range, we normalize the singular values and
set the inputs to the hypernetwork as
\begin{equation} \label{eq:gin_FNN}
 g_{\rm in} = {\left\{ \widetilde{\qsigma}, \sqrt{{\rm SNR}} \right\}},
\end{equation}
where ${\widetilde{\qsigma} = [\widetilde{\sigma}_1, \ldots, \widetilde{\sigma}_N]}$ with ${\widetilde{\sigma}_n = \sigma_n/\|\qsigma\|_2 }$ for $n=1, \cdots N$.
We can interpret $\widetilde{\qsigma}$ as a distribution shape of $\qA$ and $\sqrt{{\rm SNR}}$ as the working point of GEC-SR-Net.

Taking \eqref{eq:gin_FNN} into an input vector $\qs \in \bbR^{N+1}$,
we built our first hypernetwork, called GEC-SR-HyperNet, which is a simple two-layer network:
\begin{equation}
\qbeta ={\tt S}\left( \qW_2 \times {\tt RELU}\left(\qW_1\qs\right)\right),
\end{equation}
where ${{\tt RELU}(a) = \max(0,a)}$ and ${{\tt S}(a) = 1/\left( 1+\exp^{-a}\right)}$ are element-wise operations; and
$\qW_1 \in \bbR^{d \times (N+1)}$, $\qW_2 \in \bbR^{T \times d}$ are learnable parameters with $d$ being the size of the hidden layer in the hypernetwork.
The final layer of the hypernetwork is the sigmoid function, ${\tt S}(\cdot)$, which constrains the output in the range of damping factors (i.e., $\beta(t) \in [0, 1]$).
The learnable parameters are $\{ \qW_1, \qW_2\}$ together with $\qbeta$.
During testing, the network simply takes the input, $\qs$, to produce $\qbeta$ for GEC-SR-Net.

The above architecture uses fully connected layers, which assume that GEC-SR-Net consists of a fixed number of layers $T$.
In practice, the iteration numbers in GEC-SR should vary according to different scenarios.
If the number of layers changes, then the whole network must be retrained, which is inconvenient.

\subsubsection*{Dynamic Hypernetwork}

To enable GEC-SR-HyperNet to work with varying layer numbers,
we use a RNN to dynamically generate a damping factor that can vary online across layers.
The RNN can be in long short-term memory (LSTM) \cite{LSTM} or gate recurrent unit network (GRU) \cite{GRU} architectures.
In this work, we select GRU because it has a simple architecture and comparable performance to the LSTM.
When a hypernetwork consists of GRU, we call it HyperGRU.
We call the architecture that uses GRU as a hypernetwork to generate the damping factors for GEC-SR-Net as GEC-SR-HyperGRU.

HyperGRU is made up of connecting $T$ iteration sequence GRU (Fig. \ref{fig:Fig-HyperNet}(b)).
At every iteration $t$, GRU takes the concatenated vector of input $\qs(t)$ and the previous states, $\qh(t-1)$ as its input, and then generates the output of the current states, $\qh(t)$, and the damping factor, $\beta(t)$.
The GRU of each iteration shares the same parameters, that is, $(\qW_{\rm r},\qW_{\rm z},\qW_{\rm h},\qW_{\rm o})$. We use a standard formulation of the GRU model (Fig. \ref{fig:Fig-HyperNet}(c)), which is specifically given by
\begin{subequations}
\begin{align}
\qz(t) &= {\tt S}\left(\qW_{\rm z} [\qh(t-1),\qs(t)] \right),\\
\qr(t) &= {\tt S}\left( \qW_{\rm r} [\qh(t-1),\qs(t)]\right),\\
\widetilde{\qh}(t) &=\tanh \left( \qW_{\rm h} [ \qr(t) \odot \qh(t-1), \qs(t) ]\right),\\
\qh(t) &= \left( 1 - \qz(t) \right)\odot \qh(t-1) + \qz(t) \odot\widetilde{\qh}(t), \label{eq: combine}\\
\beta(t) &={\tt S}\left(\qW_{\rm o} \qh(t)\right),
\end{align}
\end{subequations}
where $\qr(t)$, $\qz(t)$, and $\widetilde{\qh}(t)$ represent the reset gate, update gate, and state candidate for the $t$-th iteration layer, respectively.
The $\tanh$ function is an element-wise operation defined as $\tanh(a) = \left(  e^a-e^{-a} \right) / \left(e^a+e^{-a}\right)$. 
The current states, $\qh(t)$, are computed by \eqref{eq: combine} with $\widetilde{\qh}(t)$ and ${(t-1)}$-th state information $\qh(t-1)$, where $\qz(t)$ can be regarded as the rate of a trade-off between the state candidate and the previous state.

Notably, unlike GEC-SR-HyperNet, which generates all the damping factors at a time, GEC-SR-HyperGRU generates a damping factor sequentially.
GEC-SR-HyperGRU generates either $\beta_{z}(t)$ or $\beta_{x}(t)$ at every iteration step $t$ depending on its position
by using
\begin{equation} \label{eq:gin_GRU}
 \qs(t) = {\left[ \widetilde{\qsigma}, \sqrt{{\rm SNR}}, \beta{(t-1)}, \beta{(t-2)}, v(t)  \right]^{T}},
\end{equation}
as its input, where $\beta{(t-1)}$, $\beta{(t-2)}$, and $v(t)$ are either associated with the parameters for Module A or C. The dimension of vector $\qs(t)$ is $N+4$.
More concretely, after Module A, the input parameters are $\beta_{z}{(t-1)}$, $\beta_{z}{(t-2)}$, and $v_{2z}(t)$, and the output of HyperGRU is $\beta_{z}(t)$,
while after Module C, the input parameters are $\beta_{x}{(t-1)}$, $\beta_{x}{(t-2)}$, and $v_{2x}(t)$, and the output of HyperGRU is $\beta_{x}(t)$.
In GEC-SR-HyperGRU, the input, $(\widetilde{\qsigma}, \sqrt{{\rm SNR}})$, plays the same rule as in GEC-SR-HyperNet, which dominates the essential convergence properties.
The input, $(\beta(t-1), \beta{(t-2)})$, enables the hypernetwork to refer to the previous damping factor.
Meanwhile, the input, $v(t-1)$, enables the hypernetwork to refer to the MSE state of GEC-SR-Net to generate a damping factor of GEC-SR-Net at the next layer.
Naturally, if $v(t)$ increases incrementally, then one has to reduce the damping factor to prevent divergence, but if $v(t)$ decreases slowly, then one has to decrease the damping factor to accelerate convergence speed.
Therefore, the MSE information from the GEC-SR-Net enables the hypernetwork to act as a convergence control engine.

\begin{figure}
\begin{center}
\resizebox{3.6in}{!}{
\includegraphics*{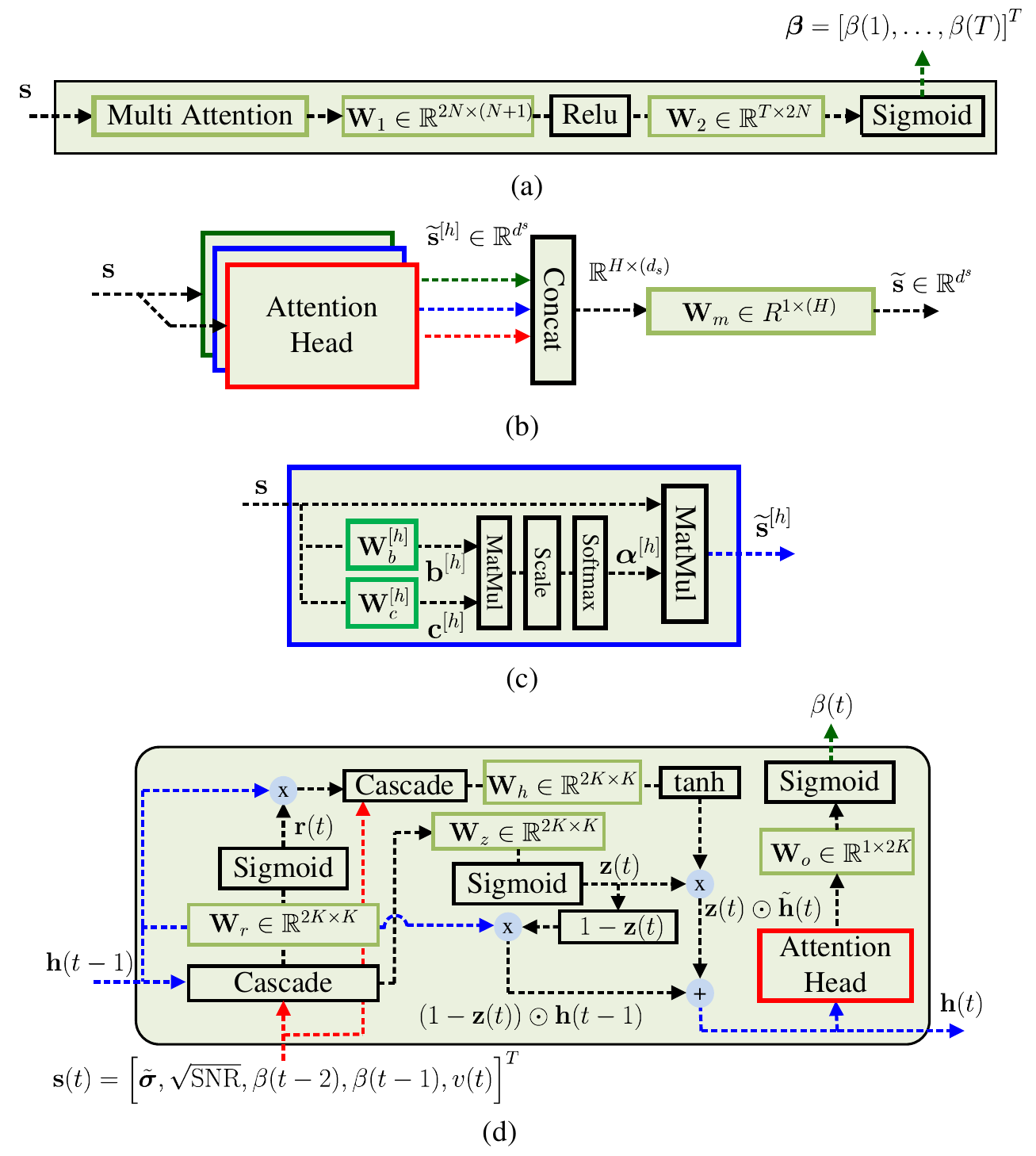} }
\caption{ Block diagrams of (a) GEC-SR-HyperNet with a multi-attention, (b) multi-attention model, (c) $h$-th attention head model, and (d) $t$-th iteration GRU model with an attention head.
}\label{fig:Fig-HyperNet-Attention}
\end{center}\vspace{-0.5cm}
\end{figure}

\subsubsection*{Hypernetworks with Attention}

The empirical experiments show that GEC-SR-HyperNet and GEC-SR-HyperGRU do not perform as well as expected when a significant mismatch in terms of statistical property occurs between the training and testing datasets.
The results are not quite reasonable because the hypernetwork should learn to adopt clinical settings and
generate the optimal damping factors.
A problem was observed from the experiments, in which the learned damping factors are only slightly changed over different scenarios.
Therefore, we infer that a good hypernetwork should be able to pay attention to different input features under various scenarios to compute a proper representation of the inputs.
Toward this end, we introduce an attention mechanism called self-attention \cite{Attention} to compute a representation of inputs.

The self-attention function, also known as intra-attention, relates the different positions of an input vector to compute a new representation of the input vector.
Let $\qs = [s_1, s_2, \cdots, s_{d_{\rm s}}]^T$ be an input vector of the attention function with dimension $d_{\rm s}$.
As illustrated in Fig. \ref{fig:Fig-HyperNet-Attention}(c), the input vector is first transformed into two feature spaces, $\qb^{[h]}$ and $\qc^{[h]}$, to calculate the attention weights, where $\qb^{[h]} = \qW_{{\rm b}}^{[h]} \qs $ and $\qc^{[h]}  = \qW_{{\rm c}}^{[h]} \qs$, with $\qW_{{\rm b}}^{[h]}, \qW_{{\rm c}}^{[h]} \in \bbR^{d_{\rm s} \times d_{\rm s}}$
being the learned parameter matrices.
Then, the output of an attention head $h$ is computed as a weighted sum
of the input elements
\begin{equation} \label{eq: h_attention}
  \widetilde{s}_{i}^{[h]} = \sum_{j=1}^{d_{\rm s}} \alpha_{ij}^{[h]} s_j,
\end{equation}
where weight coefficient $\alpha_{ij}^{[h]}$ is computed using a softmax function
\begin{equation} \label{eq:weight_coeff}
 \alpha_{ij}^{[h]} = \frac{\exp{\left( b_i^{[h]} c_j^{[h]}/ \sqrt{d_{\rm s}} \right)}}{\sum_{k=1}^{d_{\rm s}} \exp{\left( b_i^{[h]} c_k^{[h]} / \sqrt{d_{\rm s}} \right)}}.
\end{equation}
One can infer from \eqref{eq: h_attention} that $\alpha_{ij}^{[h]}$ measures the impact of the $i$-th feature on the $j$-th feature, which helps boost feature discriminability under different scenarios.

In summary, the self-attention function (or layer) can be understood as mapping input features $\qs$ into a new representation ${\widetilde{\qs}^{[h]} = [\widetilde{s}_{1}^{[h]}, \widetilde{s}_{2}^{[h]}, \cdots, \widetilde{s}_{d_{\rm s}}^{[h]} ]^T}$ that can enhance its representation capability under different scenarios.
The parameters $(\qW_{{\rm b}}^{[h]}, \qW_{{\rm c}}^{[h]})$ are different per attention head $h$.
That is, each attention head observes input features with different views, so multi-attention heads combine the attention features of each attention head
\eqref{eq: h_attention} to get $H$ attention features $[\widetilde{\qs}^{[1]}, \widetilde{\qs}^{[2]}, \cdots, \widetilde{\qs}^{[H]}]$. Eventually, we obtain a mixed feature
\begin{equation}
 \widetilde{\qs} = \sum_{h=1}^{H} w_{{\rm m}}^{[h]} \widetilde{\qs}^{[h]} 
\end{equation}
that combines different attention of views. Here, the combining weight vector,
${\qw_{{\rm m}} = [w_{{\rm m}}^{[1]}, w_{{\rm m}}^{[2]}, \cdots, w_{{\rm m}}^{[H]} ]}$, is the learned parameters.

\begin{figure}
\begin{center}
\resizebox{3.0in}{!}{
\includegraphics*{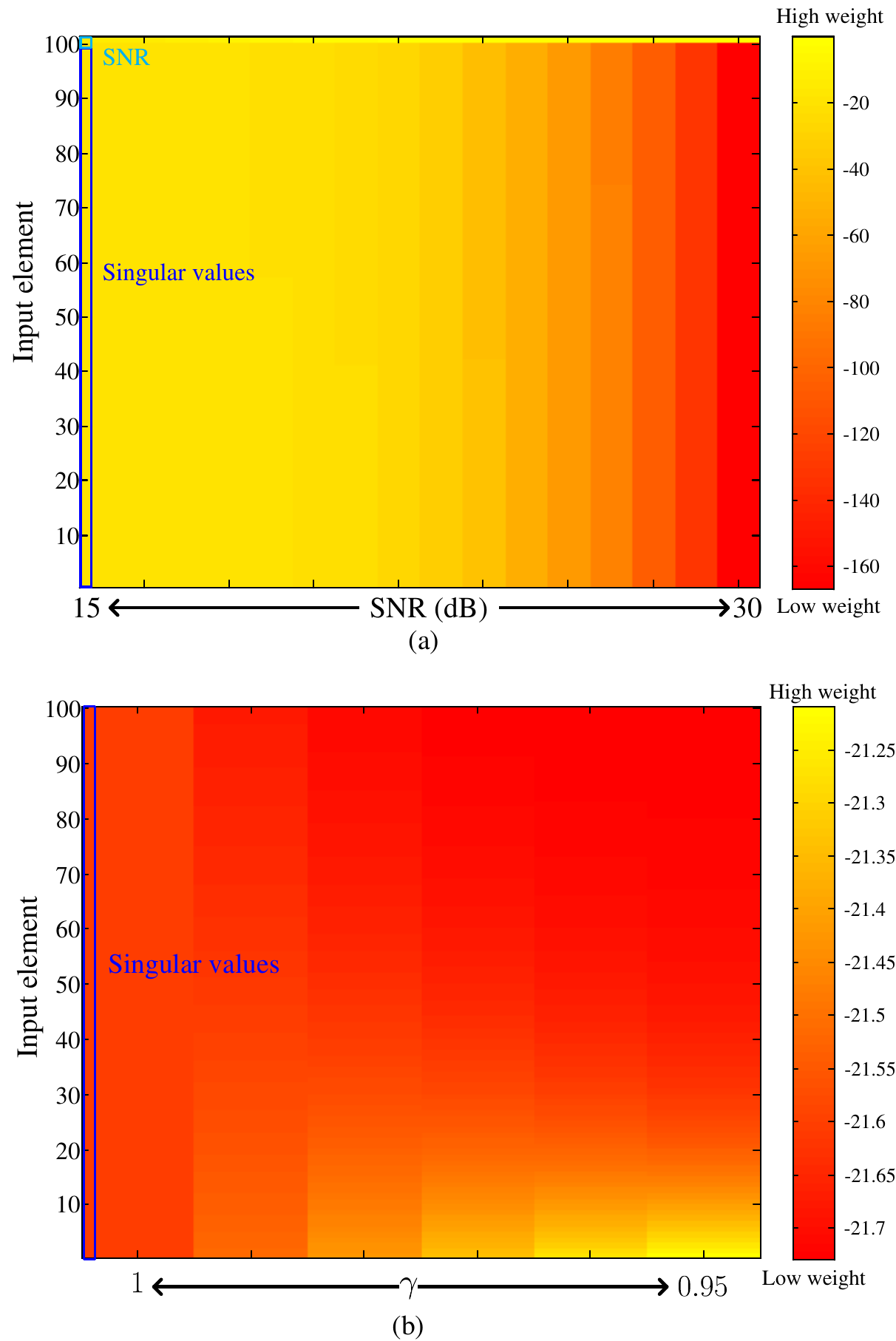} }
\caption{ Attention weights of input features in a logarithmic scale for (a) different SNR and (b) different singular distribution.
}\label{fig:Fig-Attention-weights}
\end{center}\vspace{-0.5cm}
\end{figure}

To understand the attention mechanism better, let us visualize some of the attention weights under different scenarios, whose simulation details can be referred to in Section IV.
In the examples, the input is $\qs = [\widetilde{\qsigma}, \sqrt{{\rm SNR}}]^T$, where ${\widetilde{\qsigma} \in \bbR^{100}}$ is the distribution shape of $\qA$, and the dimension of $\qs$ is $d_{\rm s} = 101$.
We use $\gamma = \widetilde{\sigma}_{n+1}/\widetilde{\sigma}_{n}$ to change the distribution shape of $\qA$.
Fig. \ref{fig:Fig-Attention-weights}(a) illustrates the weights, $[ \alpha_{11}^{[h]}, \alpha_{12}^{[h]}, \cdots, \alpha_{1 d_{\rm s}}^{[h]} ]$, of an attention head under a fixed distribution shape $\gamma = 0.98$ but
different SNRs, where the first $100$ weights are with respect to (w.r.t.) the distribution shape $\widetilde{\qsigma}$, and the last weight is w.r.t. $\sqrt{{\rm SNR}}$.\footnote{Here, we only show attention weights of an output. For the other output elements, their attention weights also have similar trend.} In the low SNR regime, the weights on
$\widetilde{\qsigma}$ and $ \sqrt{{\rm SNR}} $ are similar. The attention map increases its focus on $\sqrt{{\rm SNR}}$ as SNR rises.

Fig. \ref{fig:Fig-Attention-weights}(b) shows the weights of an attention head under fixed SNR$=20$ dB but
different distribution shapes. To facilitate the observation of effects on the different distributions of $\widetilde{\qsigma}$, we do not display the weights w.r.t. different SNRs in Fig. \ref{fig:Fig-Attention-weights}(b) because
their values remain similar under various distribution shapes.
We find that the weights w.r.t. $\widetilde{\qsigma}$ change according to the distribution shapes.
When the distribution shape is uniform (i.e., $\gamma=1$), the weights on singular values are equal.
When the distribution shape is skewed right (i.e., $\gamma=0.99 \rightarrow 0.95$), the weights focus on the singular values with large magnitude. From the examples above, for different scenarios, the weights assigned to the input features are different, and we can see the relationship between the input and output that the model intends to draw.

In GEC-SR-HyperNet, we install the multi-head attention layer in the front of the static hypernetwork (Fig. \ref{fig:Fig-HyperNet-Attention}(a)).
In GEC-SR-HyperGRU, we install a single-attention head layer after the current state $\qh(t)$ (Fig. \ref{fig:Fig-HyperNet-Attention}(d)),
because the current state, which has been reset and updated, now served as the input to generate the damping factor.
In addition, we only consider the use of single-head attention rather than the multi-head attention in HyperGRU
because the GRU model has merged input features and previous state information into the current state information.

\section*{IV. Numerical Results}

We conduct simulations to compare the GEC-SR-Net and the proposed GEC-SR-HyperNet in terms of reconstruction accuracy and convergence speed. The following default parameters have been considered.
The dimensions of transform matrix $\qA$ have ${(M,N) = (400,100)}$.
The training and testing sets contain 4,800 and 1,200 samples, respectively, with the format of $( \qy^{l}, \qA^{l}, \qx^{l} )$ per sample.
The elements of $\qx^{l}$ are generated by i.i.d. Bernoulli-Gaussian (BG) distribution\footnote{Practical signals can be better approximated by the Gaussian-mixture distribution \cite{Christopher2006}.
It has some tunable parameters that can be efficiently determined by the expectation-maximization algorithm for approximating practical signals.
The BG distribution is the simplest version of the Gaussian-mixture distribution with the fewest tunable parameters.
The proposed algorithms can be directly applied to the cases with other sophisticated prior distributions without difficulty obtaining the better reconstruction performance for images.
To clearly demonstrate our critical idea, we use the simplest BG distribution and discuss the robustness of the concerned algorithms under various mismatched cases.
The BG distribution has also been widely used in the PR literature \cite{Schniter-TSP15,Rajaei-IPOL17,Sharma-TCI19,Wang-2020,Ma-19TIT} because of its simpleness and robustness.}
\begin{equation} \label{eq:GB_distribution}
p(x) = (1-\rho) \delta (x) + \rho \mathcal{N}_{\mathbb{C}}(x ; 0, \rho^{-1})
\end{equation}
with sparsity rate $\rho$ sampled randomly from a range ${[0.3,\,0.8]}$.
The transform matrix, $\qA^{l}$, is generated according to the singular value decomposition (SVD) ${\qA^{l} = \qU \diag(\qsigma) \qV^{H}}$ format, where $\qU \in \bbC^{400 \times 400}$ and $\qV \in \bbC^{100 \times 100}$
are unitary matrices, and $\diag(\qsigma)$ is a ${400 \times 100}$ diagonal matrix with non-negative real numbers $\qsigma \in \bbR_{+}^{100}$ on the diagonal.
The unitary matrices $\qU$ and $\qV$ are drawn uniformly with respect to the Haar measure, and the singular values, $\qsigma$, are generated by two classes of distributions.
\begin{itemize}
\item The first class of $\qsigma$ is generated according to the SVD of an i.i.d. Gaussian random matrix with the element being standard complex Gaussian.
We refer to the corresponding matrix of this class as the Gaussian transform matrix.
\item The second class of $\qsigma$ is a geometric series, that is, $\gamma = \sigma_{n+1}/\sigma_{n}$, for ${ n =1, 2, \cdots, 99}$. We set $\gamma = 1$ and $0.97$ during training.
\end{itemize}
The singular values are eventually adjusted by scale level to satisfy the SNR specification defined by \eqref{eq:snr_def}, that is, $\| \qsigma \|^2  = M \times {\rm SNR}$.
We set SNR during training in the range of  $[15, \, 25]\ {\rm dB}$.
The training set consists of two classes of singular value distributions, and each class contains 2,400 samples with different SNRs, sparsity rates, and unitary matrices $\qU, \qV$. Among the training set of the second class, the sample numbers for $\gamma = 1$ and $0.97$ are equally distributed.

All the network models are implemented using Tensorflow and trained using a PC with an NVIDIA GeForce GTX 2080-Ti GPU. An Adam optimizer with a learning rate of $0.05$ and a batch size of $100$ is used to train the learnable parameters with $T=10$ layers to minimize the loss functions.

\begin{figure}
\begin{center}
\resizebox{2.5in}{!}{
\includegraphics*{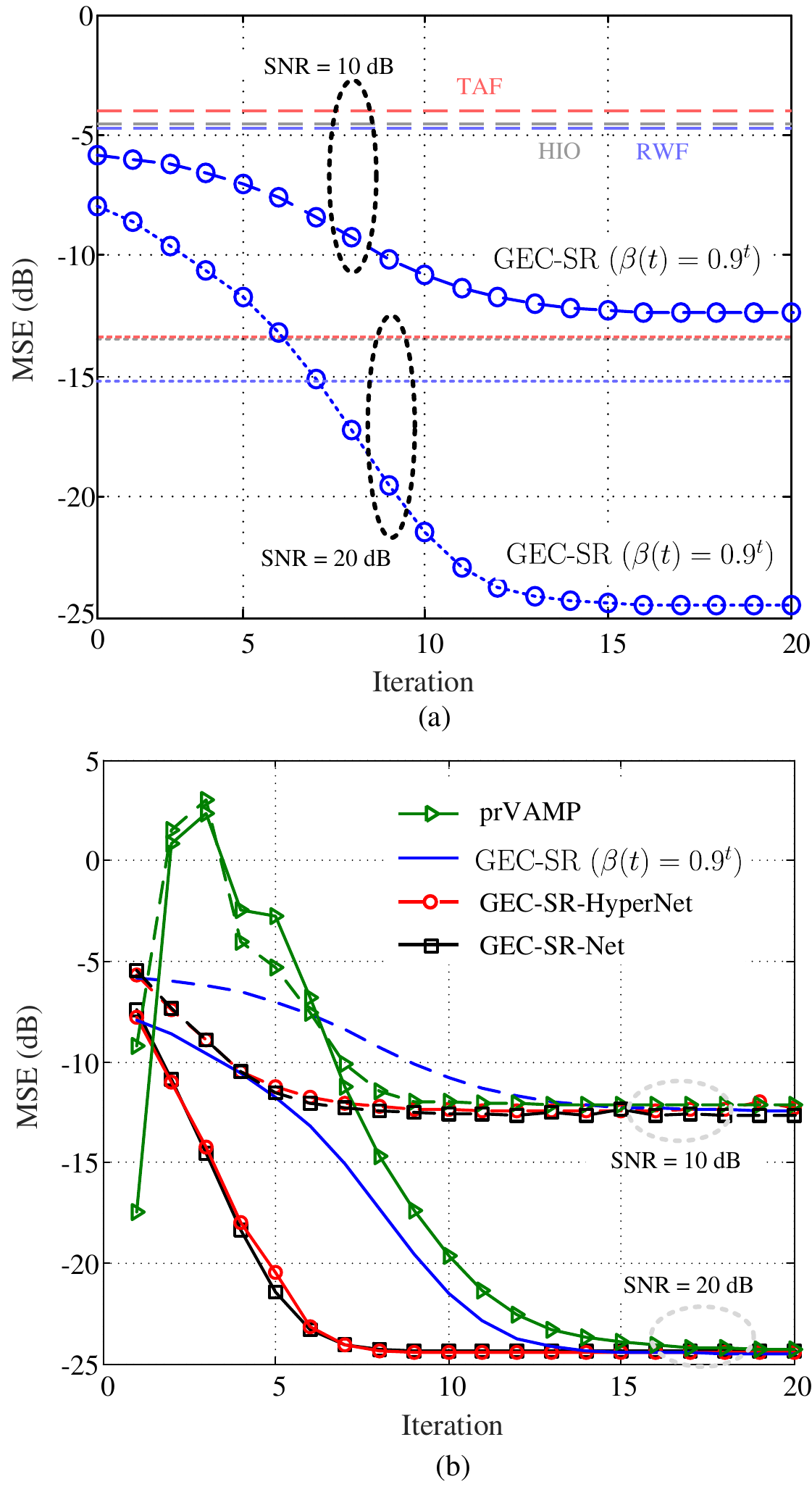} }
\caption{MSE versus the iteration for (a) the classical algorithms and (b) GEC-SR-HyperNet.
GEC-SR-Net is the model-driven learning of GEC-SR, which has the best convergence speed.
}\label{fig:Fig2-damping}
\end{center}\vspace{-0.5cm}
\end{figure}

\begin{figure}
\begin{center}
\resizebox{2.5in}{!}{
\includegraphics*{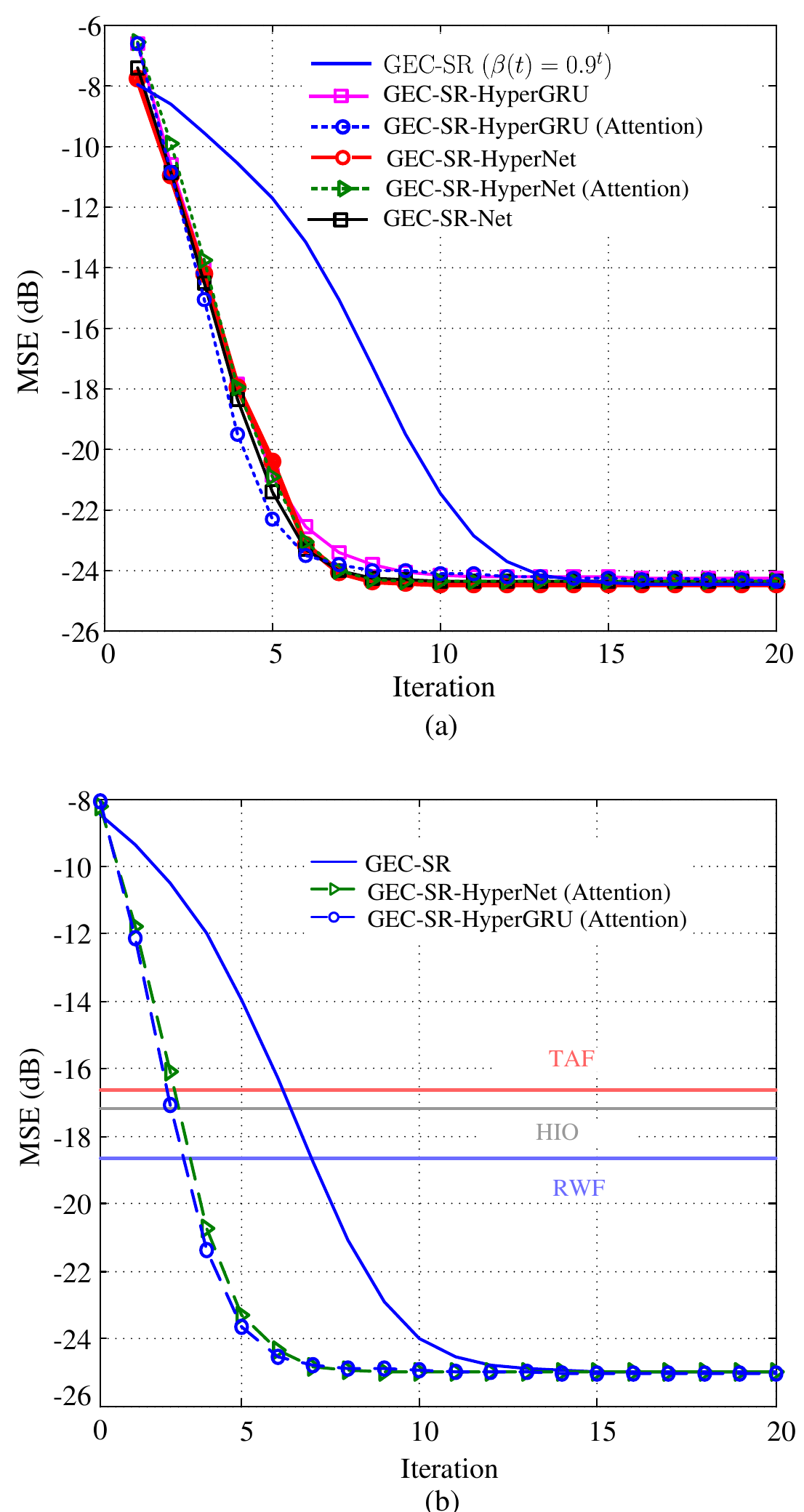} }
\caption{MSE versus the iteration for GEC-SR-Net with different hypernetworks under (a) Gaussian transform matrices and (b) Fourier transform matrices, where GEC-SR-HyperNet (Attention) uses four attention heads, and GEC-SR-HyperGRU (Attention) uses a single-attention head.}\label{fig:Fig2-damping2}
\end{center}\vspace{-0.5cm}
\end{figure}

\begin{figure*}
\begin{center}
\resizebox{7.0in}{!}{
\includegraphics*{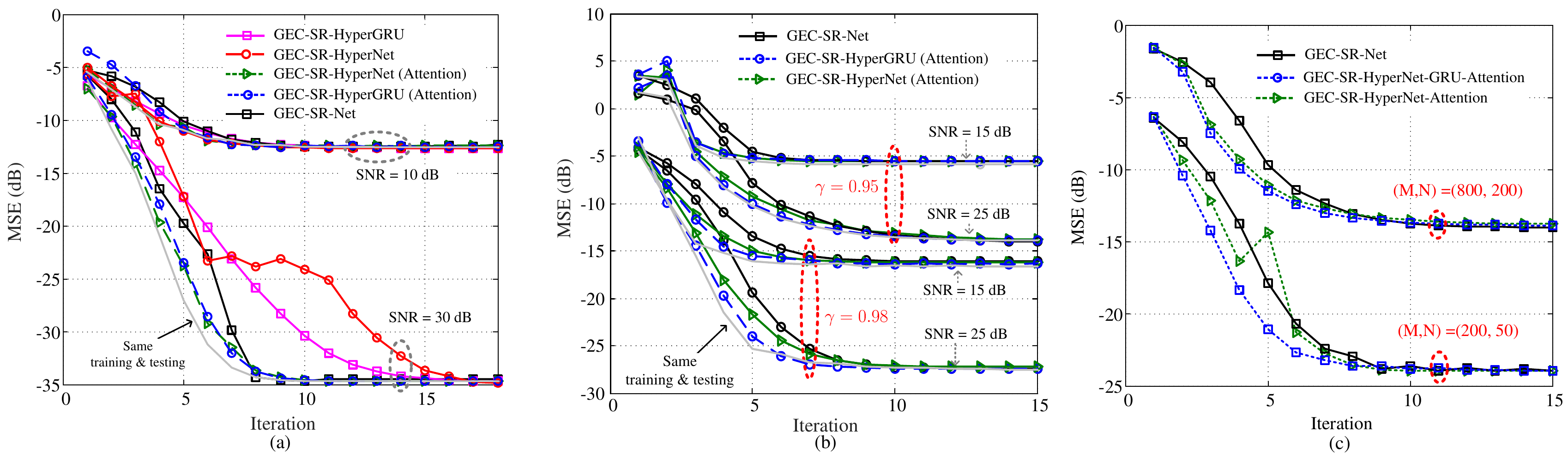} }
\caption{MSE versus the iteration for the different hypernetwork architectures under mismatched a) SNR, b) singular value distribution, and c) $N$.
GEC-SR-Net is trained under $(M,N) = (400,100)$, ${\rm SNR} = 20\ {\rm dB}$, ${\rho = 0.5}$, and the Gaussian transform matrices.
Hypernetworks  are trained under $(M,N) = (400,100)$, ${\rm SNR} = [15, \, 25]\ {\rm dB}$, ${\rho = [0.3, \, 0.8]}$, singular values with Gaussian transform matrices, and geometric series $\gamma = 1$ and $0.97$.
}\label{fig:Fig5-Robus}
\end{center}\vspace{-0.2cm}
\end{figure*}
\begin{figure*}
\begin{center}
\resizebox{7.0in}{!}{
\includegraphics*{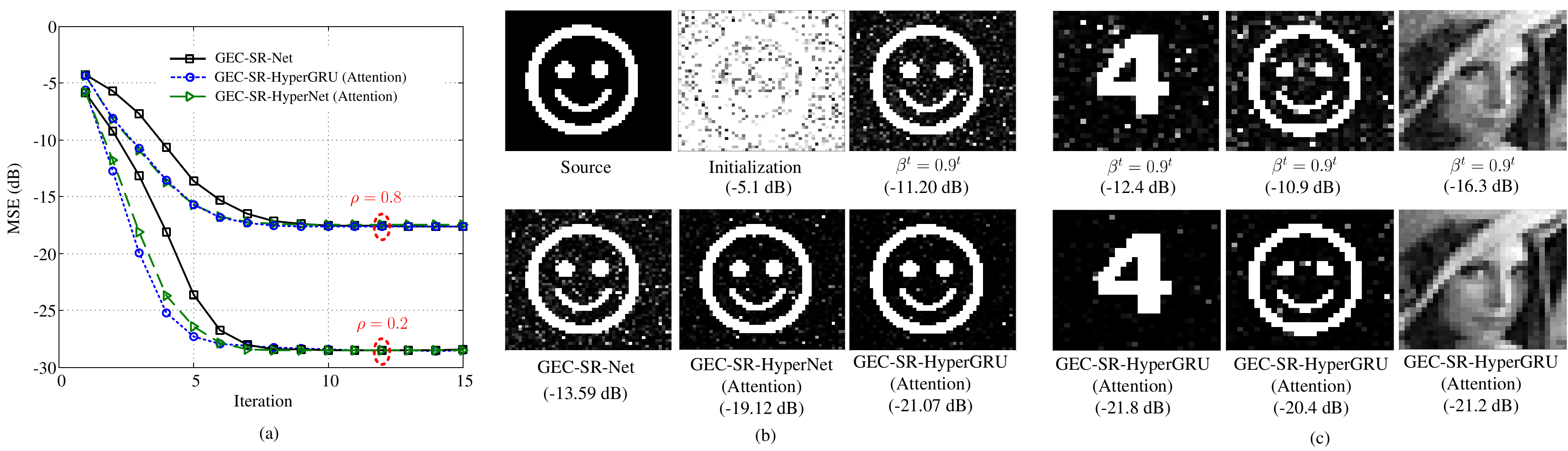} }
\caption{a) MSE versus the iteration for the hypernetworks with attention under mismatched sparsity rate $\rho$.
 b) A $50\times50$ smile face image test with different models under $t=3$.
 c) Reconstructions of three $30\times30$ images with different sparsity rates $\rho =\{0.1, 0.3, 0.9\}$ for GEC-SR and GEC-SR-HyperGRU under the ${(M,N)=(3600,900)}$ Gaussian transform matrix.
}\label{fig:Fig5-Robus2}
\end{center}\vspace{-0.2cm}
\end{figure*}
\subsection*{A. Convergence Speed}

First, we compare GEC-SR with some classical PR algorithms, such as Fienup's HIO \cite{Fienup-OL78}, RWF \cite{Zhang-17JMLR}, and TAF \cite{Wang-18TIT}, to obtain a quick sketch on the MSE and the convergence speed of the concerning algorithms. We also include prVAMP \cite{Sharma-TCI19} because it is by far the most comparable algorithm to GEC-SR. Detailed comparisons between prVAMP and GEC-SR are also available in \cite{Wang-2020}.
Fig. \ref{fig:Fig2-damping}(a) illustrates the corresponding MSE of the concerned algorithms versus the number of iterations under sparsity rate ${\rho = 0.5}$ and Gaussian transform matrix. Considering that Fienup's HIO, RWF, and TAF all need hundreds to thousands of iterations to converge, which is far beyond the scale of GEC-SR, we only indicate their convergence values in Fig. \ref{fig:Fig2-damping}(a). Clearly, GEC-SR overwhelms the competitors under various SNR scenarios.

Next, we compare GEC-SR-HyperNet with GEC-SR-Net to demonstrate the learning ability of the hypernetworks in Fig. \ref{fig:Fig2-damping}(b).
Recall that if $\beta_{x}(t) = \beta_{z}(t)$, then we simply use $\beta(t)$ to denote the damping factor.
Three damping strategies, namely, exponentially decreasing damping ${\beta(t)= 0.9^{t}}$,
\emph{learned} damping by GEC-SR-Net, and \emph{generated} damping by GEC-SR-HyperNet, are considered.
Notably, if the damping factors are manually set (e.g., ${\beta(t)= 0.9^{t}}$), then we directly refer to the reconstructor as
GEC-SR.
In Fig. \ref{fig:Fig2-damping}(b), prVAMP appears to have a divergence trajectory in iterations but converges eventually.
GEC-SR (i.e., $\beta(t) = 0.9^{t}$) shows a stable but slow convergence to an MSE level.
GEC-SR-Net and GEC-SR-HyperNet converge rapidly and are comparable because of suitable damping factors.
To obtain the benchmark performance, GEC-SR-Net is trained and tested under the datasets with the same statistical properties. For example, to test the case with ${{\rm SNR} =20}$ dB and ${\rho = 0.5}$, GEC-SR-Net is also learned to obtain the optimal damping factors under the same statistical property.
If the SNR is changed, then the damping factors of GEC-SR-Net are retrained.
As discussed before, GEC-SR-HyperNet learns to generate the optimal damping factors for each different scenario, which can also be demonstrated by comparing the convergence speed of GEC-SR-Net and GEC-SR-HyperNet.

To compare the performance of different hypernetwork architectures\footnote{The source codes to train hypernetworks and reproduce Fig. \ref{fig:Fig2-damping2}(b) are available on https://github.com/Wangchangjen/GEC-SR-PR-HyperNets},
Fig. \ref{fig:Fig2-damping2}(a) shows the convergence performance of GEC-SR-HyperNet, GEC-SR-HyperGRU, and those with attentions, under Gaussian transform matrices. From the figures, all the hypernetworks are comparable with the benchmark GEC-SR-Net.
GEC-SR-HyperGRU performs slightly better than GEC-SR-HyperNet, and the hypernetworks with attentions work slightly better than those without attentions.
Fig. \ref{fig:Fig2-damping2}(b) compares the hypernetworks with traditional algorithms under Fourier transform matrices. The hypernetworks with attentions also have considerably better performance than their competitors.

\subsection*{B. Robustness}
In previous experiments, the training and test sets are drawn from the same probability distribution, that is, they have the same singular value distributions of $\qA$, prior distributions of $\qx$, and SNR range, except Fig. \ref{fig:Fig2-damping2}(b).
In this subsection, we examine the robustness of the proposed networks under mismatched distributions. In the following experiments, GEC-SR-Net is trained under ${\rm SNR} = 20\ {\rm dB}$, ${\rho = 0.5}$, and singular values with Gaussian transform matrices. The training datasets of GEC-SR-HyperNet, GEC-SR-HyperGRU, and their associate attention models are identical to those described at the beginning of this section. When talking about robustness, we mean that no network retrains according to the test distributions.

Fig. \ref{fig:Fig5-Robus}(a) illustrates the MSE trajectories of all the concerned networks under mismatched SNRs.
In ${\rm SNR} =10$ dB, GEC-SR-HyberNet and GEC-SR-HyberGRU have faster convergence speeds than GEC-SR-Net.
However, they do not perform as well as expected at ${\rm SNR} = 30$ dB.
We analyze the generated damping factors of the two models and realize that their corresponding damping factors only slightly change between SNR$=10$\,dB and $30$\,dB. The results are not reasonable because the damping strategy should vary with SNR.
By adding the attention layers, GEC-SR-HyperNet and GEC-SR-HyberGRU can overcome this problem, and their corresponding performance improvement dramatically.
At $t=5$, GEC-SR-Net achieves ${\rm MSE} =-20$ dB, and GEC-SR-HyperNet and GEC-SR-HyberGRU with attention reach $-24$ dB.
To better understand the performance of GEC-SR-HyberGRU with attention, we also attach the best result (light-gray solid line), where the additional samples with the same SNRs are provided as testing in the training data set.
The result demonstrates that GEC-SR-HyberGRU with attention can perform close to the best results even without using the statistical properties. GEC-SR-HyberGRU with attention can be trained under a wider SNR range (i.e., $[10,\,30]$\,dB) to obtain the optimal result without any difficulty. Our training data restricted in the narrower SNR range (i.e., $[15,\,25]$\,dB) just intend to evaluate the robustness of the models. Given that the attention mechanism should always be used in GEC-SR-HyperNet and GEC-SR-HyberGRU to obtain excellent results under mismatched scenarios, we only consider GEC-SR-HyperNet and GEC-SR-HyberGRU with attentions in the following experiments. For concise expression, we simply refer to GEC-SR-HyperNet and GEC-SR-HyberGRU with attentions as GEC-SR-HyperNet and GEC-SR-HyberGRU, respectively.

Fig. \ref{fig:Fig5-Robus}(b) illustrates the corresponding MSEs of the concerned networks under mismatched singular value distributions. Again, GEC-SR-HyberGRU shows the fastest convergence speed in all the testing cases and is close to the optimal result (light-gray solid line), which is reasonable. GEC-SR-HyperNet only considers environmental features (i.e., $\qsigma$), while GEC-SR-HyperGRU considers not only the environment features but also the current state (i.e., $\qh(t)$) and convergence state (i.e., $\beta(t-2),\beta(t-1), v(t)$). In fact, GEC-SR-HyperGRU can adjust the damping factor online on the basis of its observations from the system state and convergence behavior.
 
\begin{figure*}[!htp]
\begin{center}
\resizebox{5.9in}{!}{
\includegraphics*{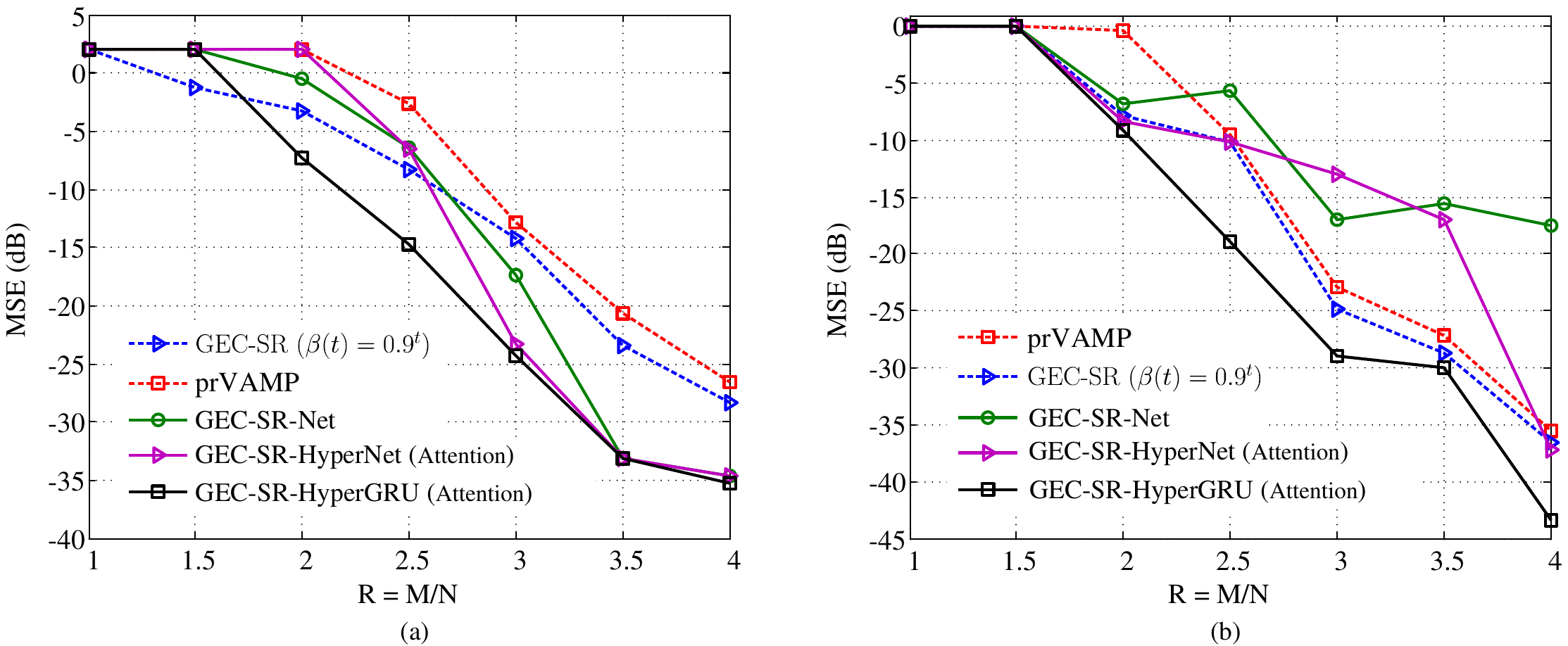} }
\caption{MSE versus the measurement ratio $R=M/N$ for the different hypernetworks of unfolded GEC-SR under a) Gaussian transform matrices and b) (0,1)-binary transform matrices.  The settings of SNR, sparsity rate $\rho$, iteration number $T$ are as follows for a) $({\rm SNR}, \rho, T)=(30\ {\rm dB},0.5,10)$ and b) $({\rm SNR}, \rho, T)=(50\ {\rm dB},1,30)$, respectively.
}\label{fig:Fig9-Measurement_ratio}
\end{center}\vspace{-0.2cm}
\end{figure*}

Fig. \ref{fig:Fig5-Robus}(c) illustrates the resulting MSE trajectories under the different dimensions of transform matrix with fixed ratio $M/N=4$ for ${\rm SNR}= 20$ dB, $\rho =0.5$, and distribution shape with $\gamma = 0.98$. We find that GEC-SR-HyperGRU also performs the best among the other network models.

Fig. \ref{fig:Fig5-Robus2}(a) illustrates the corresponding MSEs under different sparsity rates $\rho$ for ${\rm SNR}= 20$ dB, $(M,N)=(400,100)$, and $\gamma = 0.98$.
For dense prior (i.e., $\rho=0.8$), GEC-SR-HyperNet has the same performance as GEC-SR-HyperGRU, but GEC-SR-HyperGRU has better performance than GEC-SR-HyperNet for sparse prior (i.e., $\rho=0.2$).

Fig. \ref{fig:Fig5-Robus2}(b) shows the reconstructions of the $50\times50$ smile face image for different algorithms under ${\rm SNR}=15$ dB, $(M,N)=(10000,2500)$, and Gaussian transform matrix. The initialization of the image is blurry, and
the reconstruction results of all the algorithms at $t=3$ are shown. From the results, GEC-SR-HyperGRU and GEC-SR-HyperNet have the best and second-best reconstruction performance, respectively.
GEC-SR-Net only achieves $-13.59$ dB, but GEC-SR-HyperNet and GEC-SR-HyperGRU reach $-19.12$ and $-21.07$ dB, respectively. These results indicate that the hypernetworks demonstrate better robustness than GEC-SR-Net, even for real image reconstructions. To further examine the capability of GEC-SR-HyperGRU for real images,
Fig. \ref{fig:Fig5-Robus2}(c) shows the reconstruction results of three $30\times30$ images at $t=3$, $3$, $5$ and SNR $=15$, $15$, $20$\,dB and with sparsity rates $\rho =\{0.1, 0.3, 0.9\}$ and the ${(M,N)=(3600,900)}$ Gaussian transform matrix. Again, GEC-SR-HyperGRU exhibits a dramatically better performance than GEC-SR in all the tests.

GEC-SR may require matrix inversion per-iteration for certain applications. Accelerating the convergence speed also reduces the complexity significantly. To further understand the convergence speed and stability of different algorithms, Figs. \ref{fig:Fig9-Measurement_ratio}(a) and \ref{fig:Fig9-Measurement_ratio}(b) compare the algorithms under different measurement ratios $R = M/N$ with finite iterations $T$ for Gaussian and binary (i.e., the element of $\qA$ being (0,1)-binary) transform matrices, respectively.
Notably, all the networks are trained under Gaussian transform matrices with a size of $(M,N) = (400,100)$, as described at the beginning of this section.
When similar Gaussian transform matrices are used in testing, Fig. \ref{fig:Fig9-Measurement_ratio}(a) shows that GEC-SR-HyperNet has better performance than GEC-SR-Net, and GEC-SR-HyperGRU is the best among the competing algorithms. The above results demonstrate that GEC-SR-HyperGRU has a better reconstruction performance for a fixed number of iterations than the others.
For example, GEC-SR-HyperGRU can converge to $-15$\,dB for $M = 2.5N$ and $T=10$, while GEC-SR only converges to $-9$\,dB under the same condition.

In \cite{Sharma-TCI19}, the signal reconstruction becomes a challenge under binary measurement
matrices, in which many classical PR algorithms, such as prGAMP \cite{Schniter-TSP15}, PhaseMax \cite{Goldstein-TIT18}, and PhaseLamp \cite{Dhifallah-Allerton17}, show instability and fail even with high SNR.
In this regime, Fig. \ref{fig:Fig9-Measurement_ratio}(b) shows that all the GEC-SR-based algorithms (including prVAMP) also work while the number of iterations/layers are increased to $T=30$.
For GEC-SR-Net and GEC-SR-HyperNet, we set $\beta(t) = 0.5$ when $t > 10$ because we train the two models only for $T=10$ layers.
GEC-SR-Net exhibited the worst performance among the competing algorithms even under the same $M/N = 4$ setting as in training. This result is expected because
GEC-SR-Net has shown weak robustness in the previous experiments under mismatched distributions.
Moreover, GEC-SR-HyperNet cannot perform competitively because it only considers the general setting features and not the instability of the algorithms under binary measurement
matrices. In contrast to GEC-SR-HyperNet, GEC-SR-HyperGRU considers the setting features and adaptively adjusts the damping factors based on its convergence state.
Therefore, GEC-SR-HyperGRU shows stability and fast convergence speeds.

We have tested the robustness of the concerned algorithms under various mismatched distributions in terms of SNR, signal prior, transform matrix, and measurement sizes.
The results demonstrate that GEC-SR-HyperGRU has advantages
in recovery in terms of speed and accuracy. Importantly, GEC-SR-HyperGRU shows excellent robustness and retains its advantages even without retraining.

\section*{ V. Conclusion}

We have proposed a novel framework on deep unfolding.
As a practice for robust PR, we have exploited a hypernetwork in the unfolded GEC-SR, called
GEC-SR-HyperNet. GEC-SR-HyperNet incorporates the hypernetwork to generate the damping factors of GEC-SR-Net and adapt to the scenarios. 
To make the hypernetwork work with varying layer numbers, we use a RNN called GEC-SR-HyperGRU for dynamic hypernetwork. 
In addition, GEC-SR-HyperGRU considers the features of the forward model and convergence state in each iteration, thereby showing the best convergence and stability.
We also introduce the attention function to relate the different positions of the
input features under different scenarios and further improve the robustness of the hypernetwork.
Our results show that GEC-SR-HyperGRU with attention provides excellent convergence speed and stability compared with existing state-of-the-art.

When the forward operator is defined, our basis algorithm, GEC-SR, can be applied to solve other inverse problems, such as quantization.
Our evaluation under the PR setting is attributed to convergence issues, which are more challenging than quantization \cite{Wang-20ICASSP}.
Therefore, the advantages of the hypernetwork to GEC-SR for other inverse problems are expected.
Furthermore, we believe that our framework can be applied to other unfolded algorithms. For example,
the application of the hypernetwork to a PnP recovery with a DNN denoiser \cite{Ryu-arXiv19,Zhang-2020} should immediately enhance the original recovery in terms of speed, stability, and adaptability.
A policy network via obtained deep reinforcement learning is developed recently in \cite{Wei-20ICML} to determine the
internal parameters of PnP-ADMM automatically. The hypernetwork and the policy network should have their own merits, and their integration and comparison will be explored further in future works.

{\renewcommand{\baselinestretch}{1.0}
\begin{footnotesize}
\bibliographystyle{IEEEtran}

\end{footnotesize}}

\end{document}